\documentclass[preprint, authoryear, 12pt]{elsarticle}
\usepackage{amssymb}
\usepackage{natbib}
\usepackage{amsthm}
\usepackage{amsmath}
\usepackage{tabularray}
\usepackage{ulem}
\usepackage[colorlinks, citecolor=blue,linkcolor=blue]{hyperref}
\usepackage{graphicx}
\usepackage{float}
\usepackage[section]{placeins}
\usepackage{threeparttable}
\usepackage{verbatim}
\usepackage{bbding}
\usepackage{color}
\usepackage[dvipsnames]{xcolor}
\definecolor{purple}{rgb}{0, 0, 0}
\newcommand{\addtext}[1]{\textcolor{purple}{#1}}

\definecolor{light_red}{rgb}{0, 0, 0}

\newcommand{\addtexttwo}[1]{\textcolor{light_red}{#1}}
\journal{Neural Networks}

\begin{document}

\begin{frontmatter}

\title{Spiking-PhysFormer: Camera-Based Remote Photoplethysmography with Parallel Spike-driven Transformer}

\author[label1]{Mingxuan Liu \footnotemark[1]}
\author[label1]{Jiankai Tang \footnotemark[1]}
\author[label2]{Yongli Chen}
\author[label1]{Haoxiang Li}
\author[label1]{Jiahao Qi}
\author[label1]{Siwei Li}
\author[label2]{Kegang Wang}
\author[label2]{Jie Gan}
\author[label1]{Yuntao Wang \footnotemark[2]}
\author[label1]{Hong Chen \footnotemark[2]}

\affiliation[label1]{
            addressline={Tsinghua University}, 
            city={Beijing},
            country={China}}
            
\affiliation[label2]{
            addressline={Beijing Smartchip Microelectronics Technology Co., Ltd}, 
            city = {Beijing},
            country= {China}
}
            
\begin{abstract}
Artificial neural networks (ANNs) can help camera-based remote photoplethysmography (rPPG) in measuring cardiac activity and physiological signals from facial videos, such as pulse wave, heart rate and respiration rate with better accuracy. However, most existing ANN-based methods require substantial computing resources, which poses challenges for effective deployment on mobile devices. Spiking neural networks (SNNs), on the other hand, hold immense potential for energy-efficient deep learning owing to their binary and event-driven architecture. To the best of our knowledge, we are the first to introduce SNNs into the realm of rPPG, proposing a hybrid neural network (HNN) model, the Spiking-PhysFormer, aimed at reducing power consumption. Specifically, the proposed Spiking-PhyFormer consists of an ANN-based patch embedding block, SNN-based transformer blocks, and an ANN-based predictor head. First, to simplify the transformer block while preserving its capacity to aggregate local and global spatio-temporal features, we design a parallel spike transformer block to replace sequential sub-blocks. Additionally, we propose a simplified spiking self-attention mechanism that omits the value parameter without compromising the model's performance. Experiments conducted on four datasets—PURE, UBFC-rPPG, UBFC-Phys, and MMPD demonstrate that the proposed model achieves a 10.1\% reduction in power consumption compared to PhysFormer. Additionally, the power consumption of the transformer block is reduced by a factor of 12.2, while maintaining decent performance as PhysFormer and other ANN-based models.

\end{abstract}


\end{frontmatter}

\footnotetext[1]{Mingxuan Liu and Jiankai Tang are co-first authors of the article.}
\footnotetext[2]{Corresponding authors: Hong Chen and Yuntao Wang.}

\section{Introduction}
\begin{figure}[htbp]
\centering
\includegraphics[width=1\textwidth]{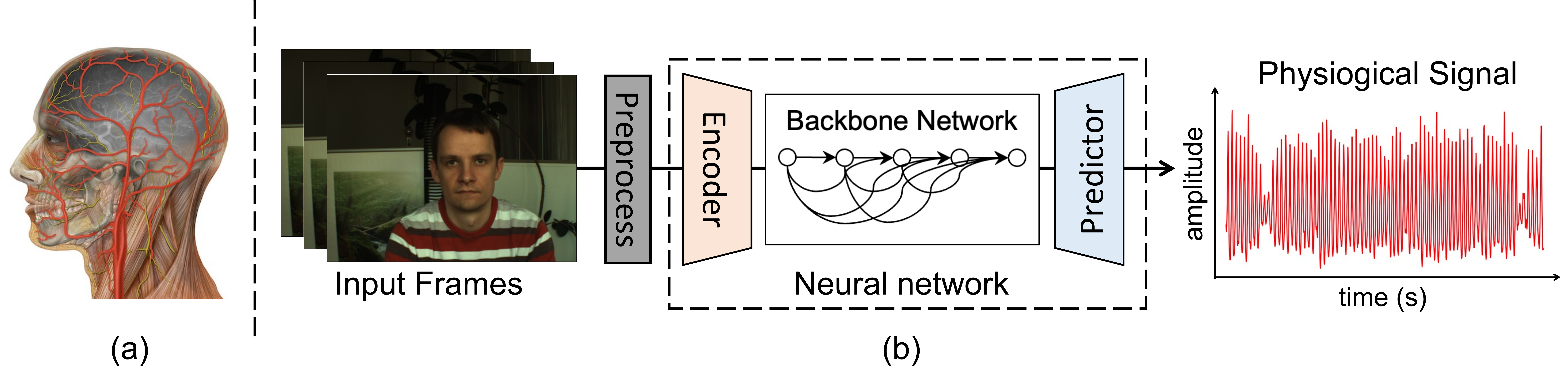}
\caption{(a) Human head anatomy with external and internal carotid arteries (\cite{headblood}) (b) rPPG pipeline of neural methods (\cite{toolbox}}
\label{fig1}
\end{figure}

The pulse wave is a vital sign of significant importance in healthcare, used to measure cardiac activity (\cite{pulsewave}). However, measuring the pulse wave via contact PPG sensors is not convenient and may not be suitable for the elderly and preterm infants (\cite{mit}). Consequently, camera-based remote photoplethysmography (rPPG) has been proposed to predict the pulse wave through light reflected off the face, thereby obtaining heart rate (HR), respiratory rate (RR), and pulse transit time (PTT) (\cite{survey}). As illustrated in Fig. \ref{fig1}(a), the periodic movement of blood from the heart to the head through the abdominal aorta and carotid artery results in periodic motion of the facial color (\cite{allen2007photoplethysmography,tra3}). The rPPG detects the pulse from the movement without contact measurements, which allows for continuous monitoring outside clinical settings, providing doctors with timely samples, offering long-term trends and statistical analysis (\cite{mit}). \textcolor{purple}{In addition, the SARS-CoV-2 (COVID-19) pandemic has increased the demand for remote diagnostics. One issue of current telemedicine systems is that physicians are not able to assess patients' physiological status remotely. With rPPG technology, the problem can be solved (\cite{toolbox}).}


Early rPPG methods used traditional signal processing to analyze facial color changes (\cite{tra1,tra2,tra3,tra4,tra5,mit,mit2}). For example, \cite{mit} tracked head features and used PCA to break down motions. \cite{mit2} introduced Eulerian Video Magnification for spatial decomposition and temporal filtering. However, these methods are limited by body movement and lighting conditions, and may be biased in the presence of motion, lighting changes, and noise (\cite{survey2}).

With the advancement of deep learning, supervised learning methods based on artificial neural networks (ANNs) (shown in Fig. \ref{fig1}(b)) have been proposed to address the aforementioned problems (\cite{aaai,deepphys,efficientphys,physformer,physformer++,physnet,eaai,tpami,bmvc}). These neural methods are mainly based on convolutional networks (CNNs) or transformers. The DeepPhys CNN by \cite{deepphys} and the two-step CNN by \cite{bmvc} are notable examples. Transformers, renowned for their exceptional long-range capabilities, have found applications in various domains such as natural language processing (NLP) and video analysis (\cite{trans1, trans2, trans5, trans6, trans7, trans8, trans9, trans13}). \cite{physformer} introduced a video transformer, and \cite{radiant} proposed RADIANT to enhance rPPG features. Despite their accuracy, ANNs, and specifically transformer models, necessitate considerable computational power. This requirement becomes particularly challenging for long-term remote photoplethysmography (rPPG) applications due to increased energy consumption. Additionally, the deployment of such models on edge computing devices is hindered by their limited memory and processing capabilities, making it difficult to support complex modules. \textcolor{purple}{Specifically, excessive computational complexity of models may render edge devices inoperable or result in high power consumption, whereas an excessive number of parameters can be constrained by limited memory.} Therefore, finding a balance between performance and computational efficiency continues to be a significant challenge in this field (\cite{single}). In response to the challenge, a new approach emerges with spiking neural networks (SNNs).

\begin{figure}[htbp]
\centering
\includegraphics[width=1\textwidth]{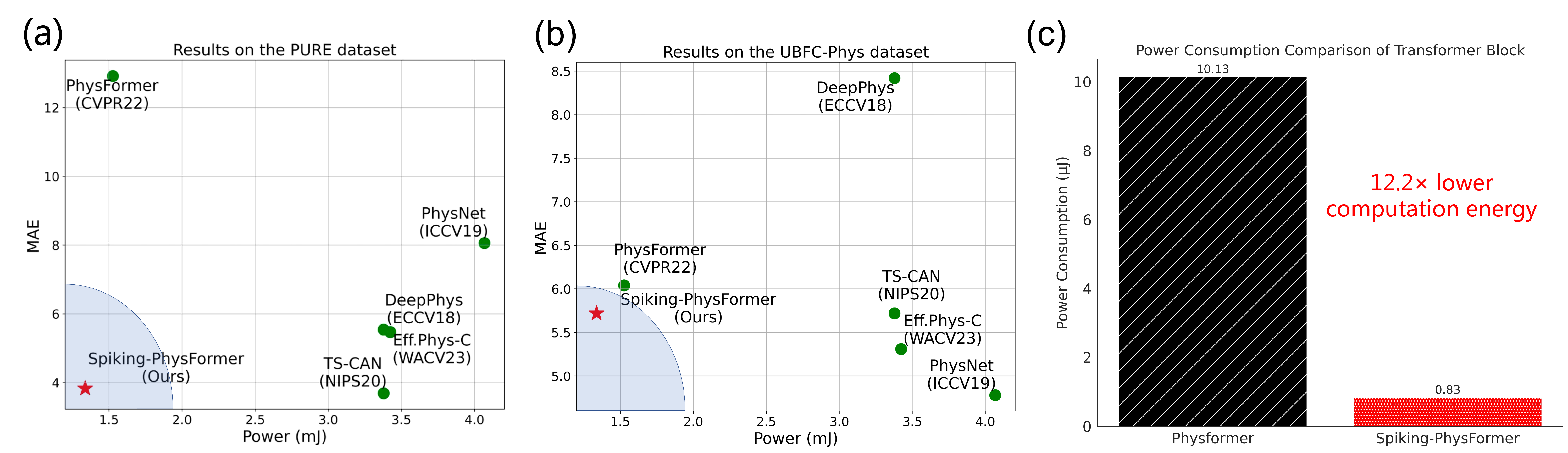}
\caption{MAE (lower is better) vs inference energy of different neural methods implemented in 45nm technology (\cite{45nm}) with an input frame size of 128 × 128, the shaded blue region shows the preferred region (The FLOPs for ANN-based models are derived from PhysBench (\cite{physbench}) and adjusted based on input size). (a) Results on the PURE dataset (\cite{pure}) after training on the UBFC-rPPG dataset (\cite{ubfc-rppg}). (b) Results on the UBFC-Phys dataset (\cite{ubfc-phys}) after training on the PURE dataset (\cite{pure}) (c) The computational energy required by the transformer block in the Spiking-PhysFormer is 12.2 times lower than that in the PhysFormer (\cite{physformer}).}
\label{fig2}
\end{figure}

As the third generation of neural networks, SNNs are designed with biological plausibility to encode and transmit information in the form of spikes, mimicking the dynamics of neurons in the brain (\cite{snn1},\cite{zhang1},\cite{zhang2},\cite{zhang4}). Compared with ANNs, SNNs' event-driven nature enables a significant reduction in energy consumption when running on neuromorphic chips (\cite{snn2}). For example, using a 28nm asynchronous SNN accelerator, SNNs achieve an inference power efficiency of 3.97 pJ/SOP and a classification accuracy of 95.7\% on the N-MNIST test dataset (\cite{snn3}). Previously, SNNs were solely utilized for simple classification tasks (\cite{snnclassification1, snnclassification2, snnclassification3}), but with the introduction of complex SNN backbones in recent years (\cite{snnres1, spike-drivenformer, spikformer,zhang3},), SNNs have been employed for intricate tasks such as image generation (\cite{spikingddim, spikingdiffusion, spikingddpm}), optical flow estimation (\cite{Optical-flow-estimation}), image deblurring (\cite{image-deblurry}), and language generation (\cite{spikegpt}). However, the potential of SNNs in rPPG has not yet been explored.

In order to reduce the computational energy used by rPPG models, we introduce SNNs into this field for the first time by proposing a new hybrid neural network (HNN) called the Spiking-PhysFormer. \textcolor{purple}{As the encoding process required by SNNs to convert feature values into spike sequences, leading to information loss during feature extraction (\cite{liu2024advancing,gerhards2023hybrid}), we utilize ANNs in the feature extraction step to improve the accuracy and employs SNNs in constructing subsequent transformer blocks. Moreover, we adopt a spike self-attention mechanism to eliminate irrelevant information for pulse wave prediction. Specifically, Spiking-PhysFormer consists of a patch embedding (PE) block, transformer blocks, and a predictor head.} To balance performance and energy efficiency, we adopt the PhysFormer design for the PE block and predictor head with ANNs, and specifically design the transformer blocks with SNNs. To enhance the transformer blocks with global spatio-temporal attention based on fine-grained temporal skin color differences, we propose a parallel spike-driven transformer, which combines temporal difference convolution (TDC) with spike-driven self-attention (SDSA) mechanisms, executing multi-layer perceptron (MLP) and attention submodules in parallel to improve efficiency while minimizing performance degradation. Additionally, we introduce simplified spiking self-attention (S3A), omitting the value parameter, further reducing the complexity of the attention sub-block. The main contributions of this paper are listed:

(1) We introduce the Spiking-PhysFormer, an HNN that integrates SNNs with transformer architecture for efficient global spatio-temporal attention in rPPG models, featuring our innovative parallel spike-driven transformer and the simplified spiking self-attention (S3A) to reduce computational complexity. Spiking-PhysFormer represents the inaugural application of SNNs in rPPG signal analysis.

(2) Experiments on four datasets—PURE (\cite{pure}), UBFC-rPPG (\cite{ubfc-rppg}), UBFC-Phys (\cite{ubfc-phys}), and MMPD (\cite{mmpd})—show that Spiking-PhysFormer cuts energy use by 10.1\% compared to PhysFormer. Its transformer block requires 12.2 times less computational energy (Fig. \ref{fig2}), while maintaining performance equivalent to PhysFormer (\cite{physformer}) and other ANN-based models.

(3) Analysis of the spatio-temporal attention map based on spike firing rate (SFR) highlights Spiking-PhysFormer's capability to effectively capture facial regions in the spatial dimension. Furthermore, it demonstrates the model's ability to identify pulse wave peaks in the temporal dimension, verifying the interpretability of the proposed method.

\section{Related work}
\subsection{Camera-based remote photoplethysmography}
Camera-based rPPG is garnering increasing research interest due to its critical importance for telemedicine and remote health monitoring. DeepPhys (\cite{deepphys}) is the first to demonstrate the superiority of deep learning over traditional signal processing algorithms such as POS (\cite{pos}) and ICA (\cite{ica}), hence current research is primarily focused on designing end-to-end models using ANNs. ANN-based methods are predominantly categorized into two types: CNN-based methods (\cite{deepphys,physnet,tscan,segppg,CDCA-rPPGNet,aaai}) and Transformer-based methods (\cite{physformer,physformer++,efficientphys,radiant,DemodulationBT}). Among these, CNNs represent the most prevalent form of supervised learning utilized for camera-based physiological measurement (\cite{survey2}). For example, a multi-task temporal shift convolutional attention network (TS-CAN) (\cite{tscan}) was introduced to predict both PPG and breathing wave signals simultaneously. Another study (\cite{segppg}) utilized a CNN-based skin segmentation network prior to signal extraction. The CDCA-rPPGNet (\cite{CDCA-rPPGNet}) incorporated an attention mechanism to fuse spatial and temporal features. To address the challenges of varying distance and head motion, two plug-and-play modules, namely the physiological signal feature extraction block (PFE) and the temporal face alignment block (TFA), were proposed (\cite{aaai}). On the other hand, transformer-based rPPG research remains relatively unexplored, with PhysFormer (\cite{physformer}) pioneering the integration of transformer blocks to amplify quasi-periodic rPPG features and refine spatio-temporal representation. PhysFormer++ (\cite{physformer++}) extended this approach with two-pathway SlowFast architecture and additional temporal difference periodic and cross-attention transformers. Furthermore, EfficientPhys (\cite{efficientphys}) adopted transformer blocks as the network backbone, eliminating the need for preprocessing steps such as face detection, segmentation, normalization, and color space transformation. Indeed, ANN-based rPPG models offer high precision but come with significant computational demands. The multi-head self-attention (MHSA) mechanism in transformer blocks, in particular, involves complex matrix multiplications (\cite{single}), posing challenges for deploying these models on edge devices.


\subsection{Transformer-based spiking neural networks}

Transformer-based ANNs have demonstrated remarkable success across various domains, such as natural language processing (NLP), computer vision, and audio processing (\cite{transsurvey, BiaS}). However, the use of self-attention (SA) mechanisms in SNNs is still in its early stage. This is primarily due to the incompatibility of traditional SA mechanisms (\cite{VSA}) with the computation characteristics of SNNs, which aim to avoid multiplicative operations and reduce computational overhead (\cite{spikformer}). Recent research focused on addressing these challenges and exploring the integration of SA in SNNs for advanced deep learning (\cite{spike-drivenformer, spikformer, yao-spiketrans, masksnn, spikegpt, spitrans1, spitrans2, spitrans3}). For instance, \cite{yao-spiketrans} proposed a temporal-wise attention SNN (TA-SNN) model to reduce redundant time steps. \cite{spitrans1} proposed a method to convert ANN-transformer into SNN, but it still relies on the conventional vanilla self-attention (VSA) mechanism. In the groundbreaking work of \cite{spikformer}, a novel spiking self-attention (SSA) mechanism was introduced, which models sparse visual features using spike-form Query, Key, and Value without the need for softmax. This marks the first transformer-based SNN. Subsequent enhancements were made to improve energy efficiency, such as the spike-driven self-attention (SDSA) in \cite{spike-drivenformer}. SDSA utilizes only mask and addition operations, completely avoiding multiplications and achieving up to 87.2 times lower computation energy compared to VSA. To further reduce energy consumption without sacrificing performance, \cite{masksnn} proposed the masked spiking transformer (MST) by incorporating the random spike masking (RSM) technique. Additionally, \cite{sparsespikformer} introduced SparseSpikformer, a co-design framework that aims to achieve sparsity in Spikformer through token and weight pruning techniques. Lastly, \cite{autost} represents the first neural architecture search (NAS) method specifically tailored for transformer-based models. The advancements have greatly enhanced the performance of SNNs, allowing them to be used in complex tasks, such as audio-visual classification, human pose tracking, and language generation (\cite{TransformerbasedSN,Wang2023SSTFormerBS,Li2023MultidimensionalAS, Zou2023EventbasedHP,Lv2023SpikeBERTAL}). In addition, the transformer-based SNNs have set the groundwork for our proposed Spiking-PhysFormer, which can capture long-range spatio-temporal attentional rPPG features from facial videos.

\subsection{Hybrid neural networks}
In recent years, there has been a growing interest in exploring the potential benefits of integrating ANNs and SNNs to achieve high-performance and low-power hybrid neural networks (HNNs). Various strategies have been investigated for combining these networks across different tasks. Specifically, SpikeGAN, proposed by \cite{snngan}, includes a conditional generator using an SNN and a discriminator using a conventional ANN. It addresses the problem of learning to emulate a spatio-temporal distribution and allows for a flexible definition of target outputs leveraging the temporal encoding nature of spiking signals. Another example is Spike-FlowNet, introduced by \cite{spike-flownet}, which enables energy-efficient optical flow estimation using sparse event camera data. Inspired by the auditory cortex, \cite{gall2023corticomorphic} proposed a CNN-SNN corticomorphic architecture, achieving auditory attention detection with an accuracy of 91.03\% based on EEG data, while maintaining a low latency. For robot place recognition, \cite{hnn-robot} proposed a multimodal hybrid neural network (MHNN) that effectively encodes and integrates multimodal cues from conventional and neuromorphic sensors. They further deployed this MHNN on the Tianjic hybrid neuromorphic chip (\cite{deng2020tianjic}) and integrated it into a quadruped robot. Furthermore, HNNs have been employed to achieve storage-efficient traffic sign recognition (\cite{zhang2023storage}) and efficient object detection in autonomous driving (\cite{efficient-objdec}). However, the application of HNNs in camera-based vitals measurement has received limited attention. To balance performance and energy efficiency in the rPPG task, we introduce the Spiking-PhysFormer. To the best of our knowledge, the proposed Spiking-PhysFormer is the first HNNs-based model in camera-based rPPG that includes a comprehensive evaluation of multiple public datasets.

\section{The proposed method}
The proposed Spiking-PhysFormer integrates transformer-based SNNs into the neural method of rPPG. We adopt the SNNs learning algorithms in SpikingJelly platform 
 (\cite{b41}). The basic computational unit is the Leaky Integrate and Fire (LIF) neuron model (\cite{b42}), which can emulate the behavior of biological neurons by generating discrete spikes and can be described by:
\begin{equation}
    H[t]=V[t-1]+\frac{1}{\tau}(X[t]-(V[t-1]-V_{reset}))
\end{equation}
\begin{equation}
    S[t]=\Theta(H[t]-V_{th})
\end{equation}
\begin{equation}
    V[t]=H[t](1-S[t])+V_{reset}S[t]
\end{equation}
where $\tau$ is the membrane time constant influencing the rate of potential change over time, $X[t]$ represents the synaptic input current at time step $t$ signifying the cumulative input from connected synapses, and $H[t]$ is the neuron's membrane potential post charging and pre-spike, derived by integrating the input current. The spike occurrence at time $t$, denoted by $S[t]$, is determined by the Heaviside step function $\Theta$, which outputs a spike (value of 1) when $H[t]$ surpasses the firing threshold $V_{th}$, indicating an action potential. The membrane potential after spiking, $V[t]$, is then updated to either remain at $H[t]$ if no spike occurs or reset to $V_{reset}$, reflecting the neuron's return to a baseline state post firing.

Above definitions and equations capture the dynamics of a LIF neural, where the membrane potential of neurons is updated based on the input current, and spikes are generated when the membrane potential surpasses a certain threshold. Due to the fact that the function $\Theta(x)$ used in Eq.(2) is non-differentiable, the surrogate gradient method is required. Specifically, we use the gradient $g'$ of the arctangent function as a replacement for $\Theta'$ in order to facilitate training of SNNs using backpropagation:
\begin{equation}
    g'(x)=\frac{\alpha}{2(1+(\frac{\pi}{2}\alpha x)^{2})}
\end{equation}

\begin{figure}[htbp]
\centering
\includegraphics[width=1\textwidth]{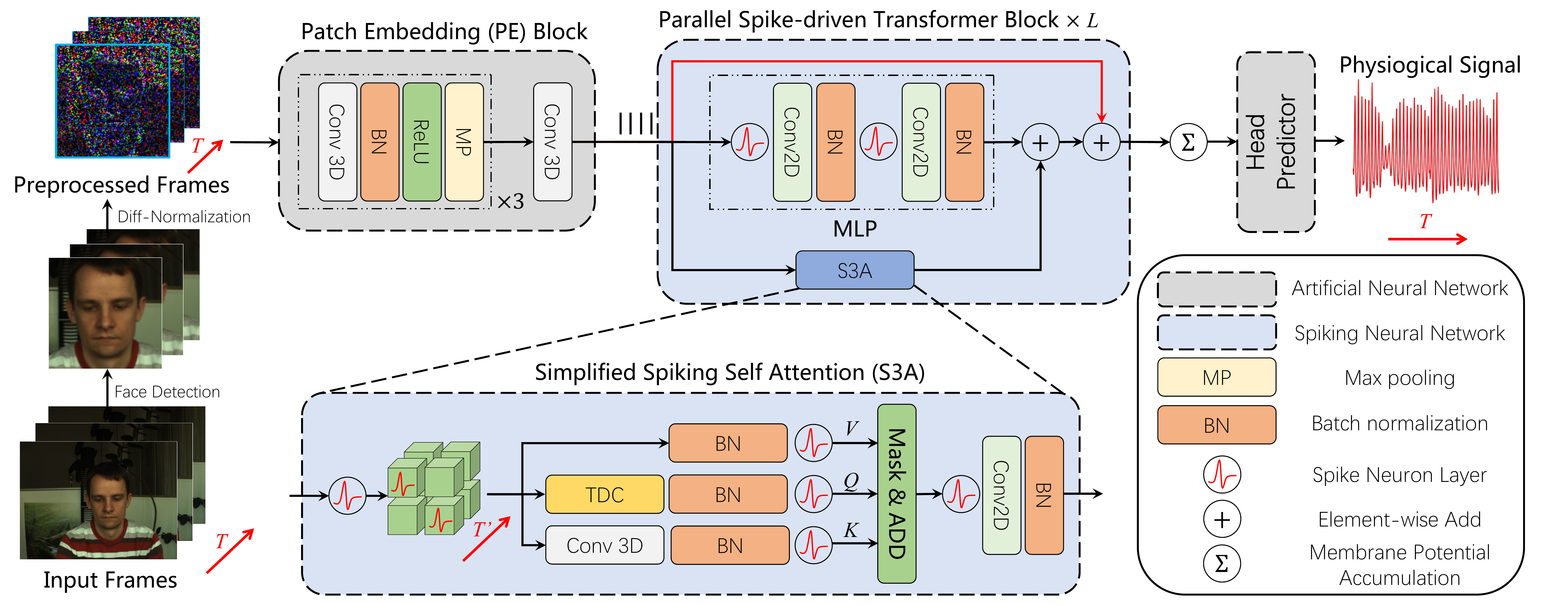}
\caption{Framework of the Spiking-PhysFormer. It consists of an ANN-based patch embedding (PE) block, several parallel spike-driven transformer blocks, and an ANN-based predictor head. \textcolor{purple}{The icon above the arrow between the PE and the Parallel spike-driven transformer blocks represents direct encoding of the output from the PE block.} For the ANN-based components of our model, we follow the network structure in PhysFormer (\cite{physformer}). Additionally, we initialize our model by pretraining PhysFormer and extracting the weights of the PE block as pre-trained parameters.}
\label{fig3}
\end{figure}

\subsection{Overall architecture}

Figure. \ref{fig3} illustrates the framework of Spiking-PhysFormer, which consists of three main components: ANN-based patch embedding (PE), parallel spike-driven transformer blocks, and an ANN-based predictor head. The PE block is utilized to extract rich spatio-temporal representations from the input video, while the simplified spiking self-attention (S3A) module in the transformer guides the model's attention towards key features. The final predictor head is responsible for mapping these features to pulse waveform peak signals.

Given an input RGB facial video, denoted as $X \in \mathbb{R}^{3 \times T \times H \times W}$, where $T$, $W$, and $H$ represent the sequence length, width, and height respectively, we begin with an initial preprocessing (IP) step to obtain preprocessed frames, denoted as $\hat{X} \in \mathbb{R}^{3 \times T \times \hat{H} \times \hat{W}}$. The PE block, comprising four 3D convolutional layers and three max pooling layers, downsampling the input frames and partitions them into spatio-temporal tube tokens $X_{tube} \in \mathbb{R}^{D \times \frac{T}{4} \times \frac{\hat{H}}{32} \times \frac{\hat{W}}{32}}$, where $D$ represents the number of channels. To prepare the input for the parallel spike-driven transformer block, we use direct encoding (DE) (\cite{b49}) to replicate $X_{tube}$ $T_s$ times, resulting in the input $U_{0}$ of shape $\mathbb{R}^{T_t \times D \times \frac{T}{4} \times \frac{\hat{H}}{32} \times \frac{\hat{W}}{32}}$. Therefore, the PE block is written as:
\begin{equation}
\begin{aligned}
    \hat{X} = {\rm{IP}}(X)
\end{aligned}
\quad
\begin{aligned}
X \in \mathbb{R}^{3 \times T \times H \times W}, \hat{X} \in \mathbb{R}^{3 \times T \times \hat{H} \times \hat{W}}   
\end{aligned}
\end{equation}
\begin{equation}
\begin{aligned}
    X_{tube} = {\rm{PE}}(\hat{X})
\end{aligned}
\quad
\begin{aligned}
X_{tube} \in \mathbb{R}^{D \times \frac{T}{4} \times \frac{\hat{H}}{32} \times \frac{\hat{W}}{32}} 
\end{aligned}
\end{equation}
\begin{equation}
\begin{aligned}
    U_0 = {\rm{DE}}(X_{tube})
\end{aligned}
\quad
\begin{aligned}
U_0 \in \mathbb{R}^{T_s \times D \times \frac{T}{4} \times \frac{\hat{H}}{32} \times \frac{\hat{W}}{32}} 
\end{aligned}
\end{equation}
Subsequently, a LIF spike neuron (SN) is used to convert $U_{0}$ into $S_0$. The spike sequence $S_0$ is then passed to the parallel spike-driven transformer blocks, which consist of a simplified spiking self-attention (S3A) block and an MLP block. As the main component in Spiking-PhysFormer, S3A offers an efficient method to model the local-global information of videos using spike-form Query ($Q$), Key ($K$), and Value ($V$) without softmax. The outputs of the MLP and S3A blocks are summed together, and the sum is then added to the input again using residual connections. After $L$ transformer blocks, the final output membrane potentials $U_{L}$ are obtained. To get a concise representation, we compute the average of $U_L$ along the spike temporal dimension, resulting in $U_{mean} \in \mathbb{R}^{D \times \frac{T}{4} \times \frac{\hat{H}}{32} \times \frac{\hat{W}}{32}}$. Finally, we utilize a Predictor Head (PH) with temporal dimension upsampling blocks to map the features extracted by the transformer-based SNN into a 1D pulse wave $Y \in \mathbb{R}^{T}$. In summary, the output of S3A, MLP and predictor head blocks can be written as follows:
\begin{equation}
\begin{aligned}
    S_l = {\mathcal{SN}}(U_l)
\end{aligned}
\quad
\begin{aligned}
S_l \in \{0,1\}^{T_s \times D \times \frac{T}{4} \times \frac{\hat{H}}{32} \times \frac{\hat{W}}{32}}, l = 0 ... L
\end{aligned}
\end{equation}
\begin{equation}
\begin{aligned}
    U_l = {\rm{S3A}}(S_{l-1})+{\rm{MLP}}(S_{l-1})+U_{l-1}
\end{aligned}
\quad
\begin{aligned}
U_l \in \mathbb{R}^{T_s \times D \times \frac{T}{4} \times \frac{\hat{H}}{32} \times \frac{\hat{W}}{32}}, l = 1 ... L
\end{aligned}
\end{equation}
\begin{equation}
\begin{aligned}
    U_{mean} = \frac{1}{T_s} \sum_{t=1}^{T_s} U_L[t,:,:,:,:]
\end{aligned}
\quad
\begin{aligned}
U_{mean} \in \mathbb{R}^{D \times \frac{T}{4} \times \frac{\hat{H}}{32} \times \frac{\hat{W}}{32}} 
\end{aligned}
\end{equation}
\begin{equation}
\begin{aligned}
    Y = {\rm{PH}}(U_{mean})
\end{aligned}
\quad
\begin{aligned}
Y \in \mathbb{R}^{T}
\end{aligned}
\end{equation}

\subsection{Data initial preprocessing}
The proposed Spiking-PhysFormer model relies on the estimated PPG signal derived from subtle variations in facial skin color caused by the cardiac pulse cycle. Therefore, it is critical to preprocess the signal to ensure accurate face detection in each frame. In line with established techniques (\cite{toolbox}), we adopt the straightforward Haar cascade detector (\cite{haar}). Then, we utilize DiffNormalized, a method that calculates the difference between consecutive frames and labels, and then normalizes them based on their standard deviation. While some approaches, such as MSTMap (\cite{mstmap}), employ regions of interest (ROI) segmentation and landmark detection to extract crucial regions from detected faces and enhance prediction accuracy, our S3A module incorporates a spatio-temporal attention mechanism that automatically focuses on salient regions, rendering the aforementioned steps superfluous.

\subsection{Parallel spike-driven transformer}

\begin{figure}[htbp]
\centering
\includegraphics[width=1\textwidth]{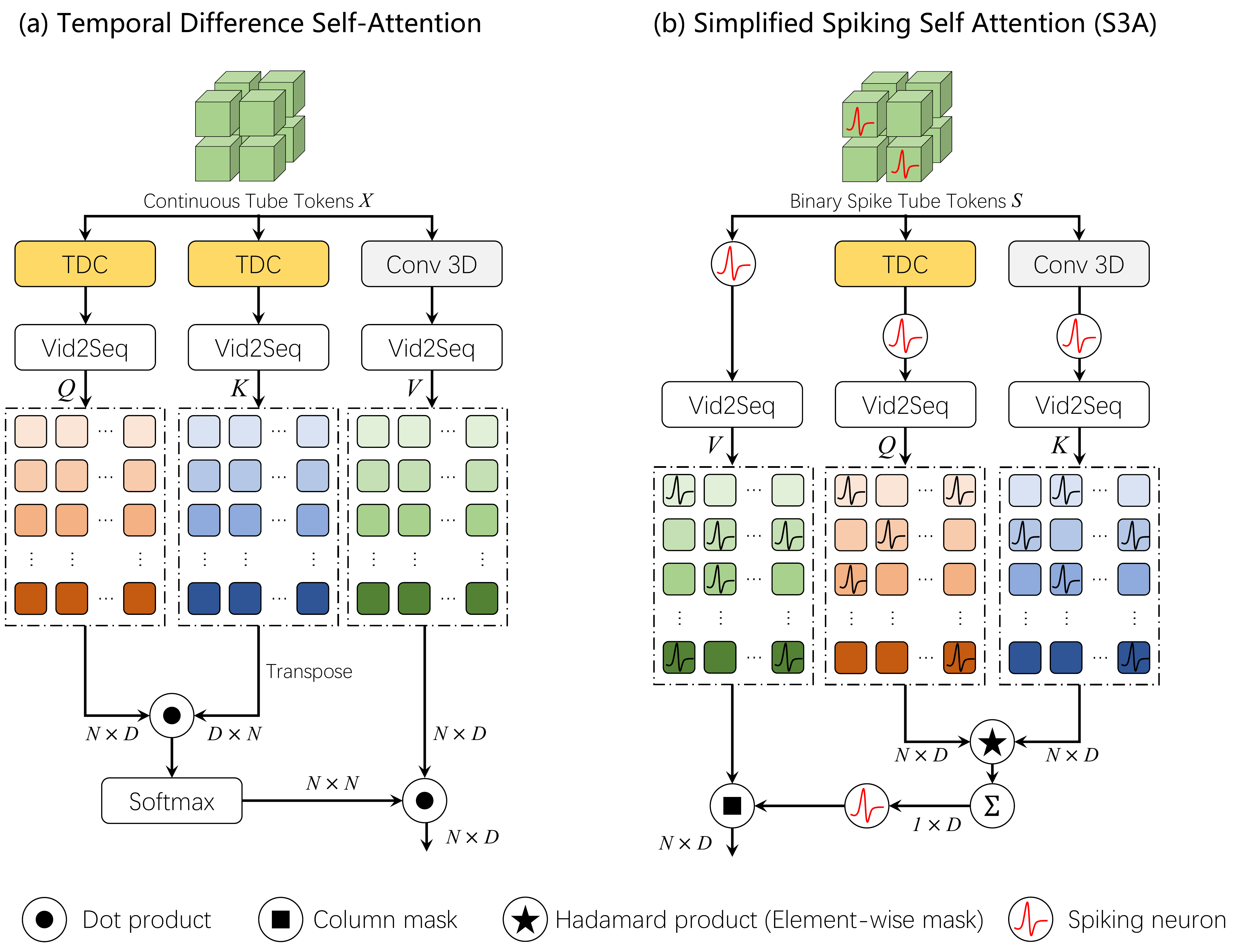}
\caption{Comparison temporal difference self-attention (TDSA) used in PhysFormer (\cite{physformer}) and our simplified spiking self-attention (S3A). (a) In TDSA, $Q$, $K$, and $V$ are obtained through linear projections using TDC (\cite{tdc}) and Conv3D. Since the input $X$ is a floating-point matrix, this involves a significant amount of multiplication operations. Furthermore, the subsequent SA operation involves matrix multiplication, specifically requiring $2N^2D$ multiply-and-accumulate operations, where $N$ is the number of tokens, $D$ is the channel dimensions. (b) Compared with TDSA, S3A utilizes TDC exclusively for query computation. Additionally, since the input $S$ is a binary spike sequence, the linear operation involved here is limited to addition. For SA computation, S3A employs an element-wise mask (Hadamard product), column summation, and column mask. As a result, only $fND$ accumulate operations are required, where $f$ represents the non-zero ratio of the matrix after applying the mask to $Q$ and $K$. Typically, $f$ is less than 0.06 (Fig. \ref{fig8}).}
\label{fig4}
\end{figure}

\textbf{Spike coding and decoding.}
To bridge the SNN block with the ANN blocks before and after SNN block, it is necessary to perform spike coding at the beginning of the transformer and decoding at the tail. The parallel spike-driven transformer blocks operate over spikes $s \in \{0, 1\}$, i.e. temporal binary values, so we perform spike coding on the features extracted by the PE block. Previous approaches employed temporal coding (\cite{timecoding}) and Poisson coding, which may bring information loss, particularly when the number of spike time steps decreases, leading to further loss of features. Although direct coding can preserve all feature representations, it brings the loss of sparsity in the computation of $Q$, $K$, and $V$ in the first transformer block. To address the problem, we introduce a fixed-parameter LIF neuron before the first S3A sub-block and MLP sub-block. The features extracted by the PE block are expanded dimensionally using direct coding and connected to the LIF neuron. This spike coding method interprets the values of the feature vectors inputted at each time step as the current intensity reaches the neuron membrane. For spike decoding, we straightforwardly average the temporal dimension of the output of the final transformer block. As the spike coding and decoding processes described above do not affect the feasibility of backpropagation, the SNN component of Spiking-PhysFormer can be trained using Surrogate Gradient (SG) methods (\cite{snnsurvey}). The loss functions of PhysFormer (\cite{physformer}) is adopt, i.e.:
\begin{equation}
    \mathcal{L}_{\rm{overall}} = \alpha \cdot \mathcal{L}_{\rm{time}} + \beta \cdot (\mathcal{L}_{\rm{CE}}+\mathcal{L}_{\rm{LD}})
\end{equation}
\begin{equation}
    \beta = \beta_{0} \cdot \eta^{\frac{{\rm{Epoch}}_{\rm{current}}-1}{{\rm{Epoch}}_{\rm{total}}}}
\end{equation}
where $\mathcal{L}_{\rm{time}}$ represents the Negative Pearson loss, which captures temporal constraints, while $\mathcal{L}_{\rm{CE}}$ reflects frequency constraints through frequency cross-entropy loss. $\mathcal{L}_{\rm{LD}}$ is a label distribution loss used for learning the distribution in HR estimation. 

\textbf{Parallel sub-blocks.}
Previous Transformer-based SNNs (\cite{spike-drivenformer,spikformer,masksnn, spikegpt, sparsespikformer, autost}) adhered to the format of a standard transformer block, in which, the output $U_{\rm{out}}$ is derived from the input $U_{\rm{in}} \in \mathbb{R}^{T_s \times N \times D}$ containing $N$ tokens and dimension $D$ using two sequential sub-blocks (one SA and one MLP) with residual connections:
\begin{equation}
    U_{\rm{out}} = \alpha_{\rm{FF}}\hat{U}+\beta_{\rm{FF}}{\rm{MLP}}(\mathcal{SN}(\hat{U}))
\end{equation}
\begin{equation}
    \hat{U} = \alpha_{\rm{SA}}U_{\rm{in}}+\beta_{\rm{SA}}{\rm{SA}}(\mathcal{SN}(U_{\rm{in}}))
\end{equation}
where scalar gain weights $\alpha_{\rm{FF}}$, $\beta_{\rm{FF}}$, $\alpha_{\rm{SA}}$, $\beta_{\rm{SA}}$ fixed to 1 by default. In our work, to simplify the traditional transformer block, we draw inspiration from the use of parallel blocks in GPT-J-6B by \cite{gpt-j} and \cite{simplifying}, and remove the residual connections in the MLP sub-blocks, obtaining the following output:
\begin{equation}
    U_{\rm{out}} = \alpha_{comb}U_{\rm{in}} + \beta_{\rm{FF}}{\rm{MLP}}({\mathcal{SN}(U_{\rm{in}}}))+\beta_{\rm{SA}}{\rm{SA}}(\mathcal{SN}(U_{\rm{in}}))
\end{equation}
with skip gain $\alpha_{comb} = 1$, and residual gains $\beta_{\rm{FF}}=\beta_{\rm{SA}}=1$ as default. Moreover, by parallelizing the processing of SA and MLP blocks, our model saves the computation time per layer. Additionally, the attention mechanism based on SFR explicitly provides crucial features (Section 4.5), so that parallel processing does not compromise model performance.

\textbf{Simplified spiking self-attention (S3A).}
The S3A sub-block is a crucial module in Spiking-PhysFormer, designed to extract key information from the spatio-temporal features generated by the PE block. A comparison between S3A and the temporal difference self-attention (TDSA) in PhysFormer is illustrated in Fig. \ref{fig4}. To capture subtle differences in local temporal features, both S3A and TDSA utilize temporal difference convolution (TDC) (\cite{tdc}):
\begin{equation}
    \operatorname{TDC}(x)=\underbrace{\sum_{p_{n} \in \mathcal{R}} w\left(p_{n}\right) \cdot x\left(p_{0}+p_{n}\right)}_{\text {vanilla 3D convolution }}+\theta \cdot \underbrace{(-x\left(p_{0}\right) \cdot \sum_{p_{n} \in \mathcal{R}^{\prime}} w\left(p_{n}\right))}_{\text {temporal difference term }}
\end{equation}
where $w$ represents the learnable weight parameters, the variables $p_{0}$, $\mathcal{R}$, and $\mathcal{R}^{\prime}$ represent the current spatio-temporal location, the sampled local neighborhood of size $(3 \times 3 \times 3)$, and the sampled adjacent neighborhood, respectively. TDC aggregates temporal difference clues within local temporal regions, but the computational cost is high due to multiple convolution operations. Specifically, for an input $S \in \mathbb{R}^{4 \times 96 \times 40 \times 4 \times 4}$, the FLOPs of TDC are approximately 28 times that of a vanilla 3D convolution layer (the result is obtained with the tool \href{https://github.com/facebookresearch/fvcore/blob/main/docs/flop_count.md}{FlopCountAnalysis}). In addition, we have applied DiffNormalized method (\cite{deepphys}) to the input videos during the first preprocessing stage to extract inter-frame differences. Therefore, we utilize TDC exclusively for calculating the query ($Q$), and a vanilla 3D convolution layer for computing the key ($K$), leading to a significant reduction in computational cost. 

\addtexttwo{To further simplify spiking self-attention, we remove the 3D convolution layer (${\rm{W}}^V$) required for the computation of the value ($V$). This is because ${\rm{W}}^V$ is simply a linear projection of the input sequence representation $x$, and the additional capacity provided by such a matrix is not particularly substantial. Previous studies (\cite{simplifying}) have shown that the value and projection parameters in the attention mechanism contribute minimally to the model's performance. The essential operations of self-attention—calculating relationships between different parts of the input—are primarily handled by the query ($Q$) and key ($K$) matrices. By setting ${\rm{W}}^V$ to the identity matrix or removing it entirely, the model maintains virtually the same performance while reducing the number of parameters and computational overhead. This leads to faster training and inference times without sacrificing accuracy. Our ablation experiments (Section 4.6) confirm that removing these additional convolution layers does not significantly impact model performance.}

As a result, $Q$, $K$, and $V$ $\in$ $\mathbb{R}^{T_s \times N \times D}$ are projected as:
\begin{equation}
    Q = \operatorname{Vid2Seq}(\mathcal{SN}({\rm{BN}}({\rm{TDC}}(S))))
\end{equation}
\begin{equation}
    K = \operatorname{Vid2Seq}(\mathcal{SN}({\rm{BN}}({\rm{Conv3D}}(S))))
\end{equation}
\begin{equation}
    V = \operatorname{Vid2Seq}(\mathcal{SN}({\rm{BN}}(S)))
\end{equation}
To obtain $V$, we normalize the spike sequence $S$ and input it into a LIF neuron for a second excitation, which decreases the sparsity of $V$, leading to less energy needs for SA computation (Section 4.3). Upon receiving $Q$, $K$, and $V$, the SA operation can be expressed as:
\begin{equation}
    \operatorname{S3A'}(Q, K, V) = g(Q, K) \otimes V = \mathcal{SN}(\operatorname{SUM}_c(Q \otimes K)) \otimes V
\end{equation}
\begin{equation}
    \operatorname{S3A}(Q, K, V) = \operatorname{Seq2Vid}(\operatorname{BN}(\operatorname{Conv2D}(\mathcal{SN}(\operatorname{S3A'}(Q,K,V)))))
\end{equation}
where Vid2Seq and Seq2Vid respectively denote the processes of flattening a video's temporal, height, and width dimensions, and subsequently restoring them to their original dimensions. $\otimes$ represents the Hadamard product. The function $g(\cdot)$ is responsible for computing the attention map, and $\operatorname{SUM}_c(\cdot)$ signifies the column-wise summation. Both $g(\cdot)$ and $\operatorname{SUM}_c(\cdot)$ yield $D$-dimensional row vectors. The Hadamard product applied to spike tensors corresponds to the mask operation. Compared with TDSA, which necessitates $2N^2D$ multiply-and-accumulate operations, the proposed S3A framework requires much less computation because $fND$ accumulation is merely needed when computing the attention map. Here, $f$ denotes the sparsity factor, representing the proportion of non-zero elements in the matrix post-application of a mask to $Q$ and $K$, $N$ corresponds to the token count, and $D$ indicates the channel dimensions. It is noteworthy that typically, $f$ is less than 0.05, as illustrated in Fig. \ref{fig8}. \textcolor{purple}{In summary, the proposed S3A simplifies calculations relative to TDSA by primarily streamlining the computation of $Q$, $K$, and $V$, and leveraging additive operations instead of multiplicative ones in the attention mechanism to reduce computational load. Subsequent ablation experiments have demonstrated that the proposed method effectively balances performance and power consumption.}

\section{Experimental results}
\subsection{Datasets and performance metrics}
\begin{figure}[htbp]
\centering
\includegraphics[width=0.8\textwidth]{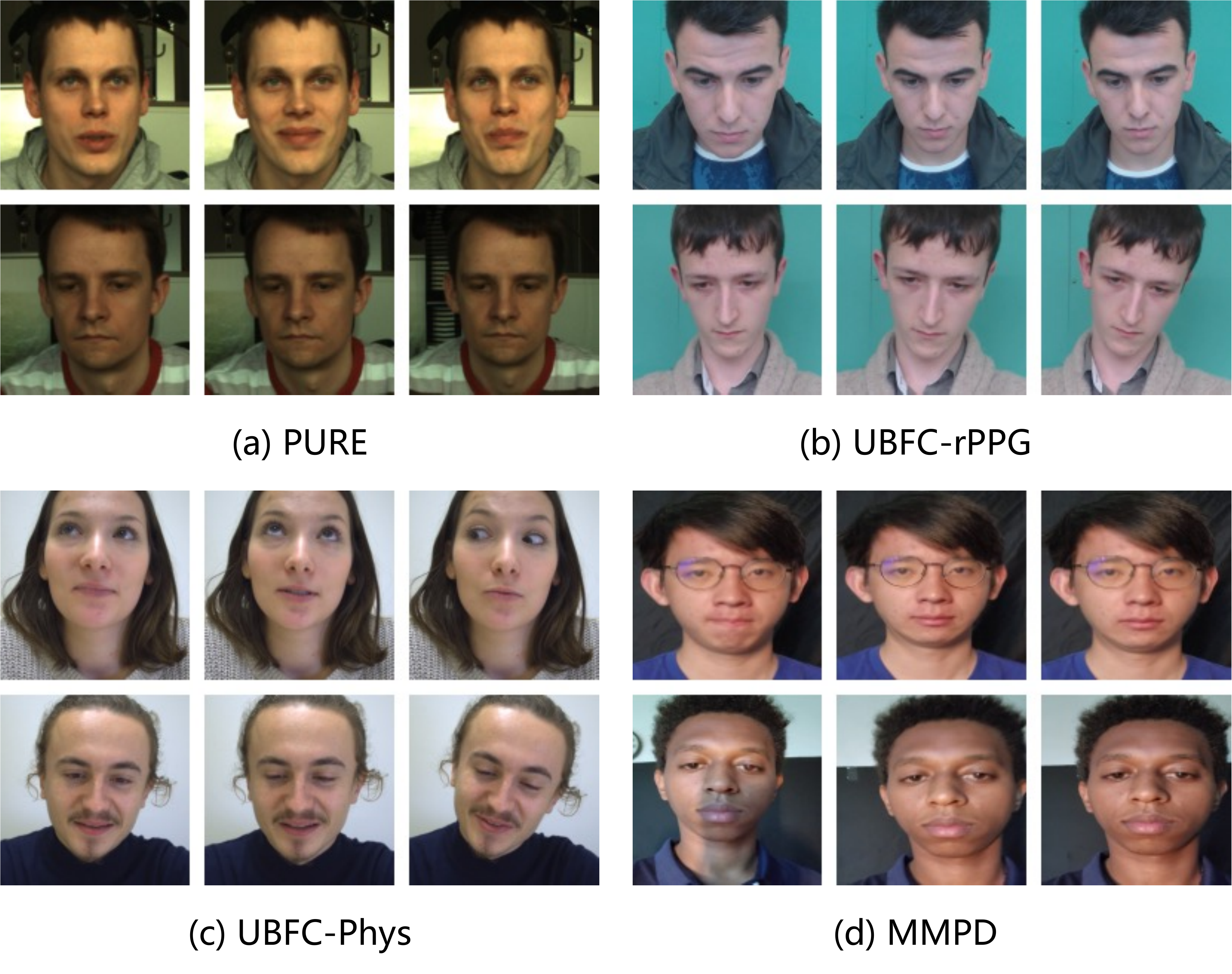}
\caption{Example video frames from datasets. (a) PURE (\cite{pure}); (b) UBFC-rPPG (\cite{ubfc-rppg}); (c) UBFC-Phys (\cite{ubfc-phys}); and (d) MMPD (\cite{mmpd}).}
\label{fig5}
\end{figure}

Experiments of rPPG-based physiological measurement for pulse wave are conducted on four benchmark datasets (PURE (\cite{pure}), UBFC-rPPG(\cite{ubfc-rppg}, UBFC-Phys(\cite{ubfc-phys}), and MMPD (\cite{mmpd}))

\textbf{PURE} (\cite{pure}): This dataset comprises vital sign measurements and video recordings from a cohort of 10 subjects, consisting of eight males and two females. Each participant engaged in six distinct recording sessions, encompassing a variety of motion conditions, thereby yielding a comprehensive dataset reflective of diverse physical states. Spatial configuration during these sessions was standardized, with subjects positioned at an approximate distance of 1.1 meters from the recording apparatus. Illumination was achieved through the strategic utilization of ambient natural light, which permeated through a window, ensuring consistent front-facing lighting conditions.  Video data acquisition was facilitated through the deployment of an RGB eco274CVGE camera, procured from SVS-Vistek GmbH, featuring operational parameters including a 30Hz frequency and a resolution specification of 640x480 pixels. For the purpose of establishing a robust gold-standard ground truth, photoplethysmography (PPG) and blood oxygen saturation (SpO2) levels were meticulously recorded at a frequency of 60Hz, employing a CMS50E pulse oximeter affixed to each subject's finger.

\textbf{UBFC-rPPG} (\cite{ubfc-rppg}):
This dataset encompasses a series of RGB video recordings from 42 subjects, derived from a scenario wherein subjects engaged in a time-limited digital game, simulating typical activities performed in front of a computer. The participants were strategically positioned approximately one meter away from the recording device during these sessions.  These sessions were conducted under indoor conditions, utilizing a blend of natural sunlight and artificial illumination to ensure optimal lighting. In terms of technical specifics, the videos were captured using a Logitech C920 HD Pro webcam, operating at a frequency of 30Hz and delivering a resolution of 640x480. These recordings are preserved in an uncompressed 8-bit RGB format. Concurrently, reference photoplethysmography (PPG) data was meticulously acquired using a CMS50E transmissive pulse oximeter, providing a gold-standard dataset for validation purposes.

\textbf{UBFC-Phys} (\cite{ubfc-phys}):
The dataset encompasses recordings from 56 subjects (46 women and 10 men), engaging in three distinct tasks characterized by considerable unconstrained motion under static lighting conditions. These tasks include a rest task, a speech task, and an arithmetic task, each designed to elicit varying physiological responses. The dataset is further enriched with gold-standard blood volume pulse (BVP) and electrodermal activity (EDA) measurements, meticulously captured using the Empatica E4 wristband. In terms of visual data, the recordings were executed using an EO-23121C RGB digital camera, ensuring high-resolution imagery at 1024x1024 pixels and a frame rate of 35Hz. For evaluation purposes, we adhered to the same subject sub-selection list and task framework as outlined in the second supplementary material of Sabour et al.~\cite{ubfc-phys}. 

\textbf{MMPD} (\cite{mmpd}): 
This dataset, meticulously assembled for comprehensive analysis, features 660 one-minute videos, showcasing a diverse array of 33 subjects (16 males and 17 females) engaging in four distinct activities (stationary, head rotation, talking, and walking) and additional exercise scenarios. The subjects, representing Fitzpatrick skin types 3-6 from multiple countries, were exposed to four different lighting conditions (LED-low, LED-high, incandescent, natural). For the purpose of validating physiological signals, ground truth photoplethysmogram (PPG) signals were concurrently captured using an HKG-07C+ oximeter at a sampling rate of 200 Hz, later downsampled to 30 Hz. The video data, concurrently recorded using a Samsung Galaxy S22 Ultra mobile phone, was initially captured at 30 frames per second with a resolution of $1280 \times 720$ pixels, before being compressed to $320 \times 240$ pixels.

\textbf{Preprocess:} \textcolor{purple}{The four datasets are processed using the pipeline described in Section 3.2 to generate video segments, each of which has the size of $128 \times 128 \times 160$, and contains 160 frames. After preprocessing, the number of videos generated for the PURE, UBFC-rPPG, UBFC-Phys, and MMPD datasets are 750, 351, 3939, and 7260, respectively.}

\textbf{Metrics}: \addtexttwo{We benchmark Spiking-PhysFormer against state-of-the-art (SOTA) ANN-based rPPG models: TS-CAN (\cite{tscan}), PhysNet (\cite{physnet}), DeepPhys (\cite{deepphys}), Eff.Phys-C (\cite{efficientphys}), PhysFormer (\cite{physformer}), iBVPNet (\cite{ibvpnet}), rFaceNet (\cite{rfacenet}), DiffPhys (\cite{DiffPhysES}), PhysNet-XY (\cite{PhysNet-XY}), and PhysNet-UV (\cite{PhysNet-XY}). For rFaceNet, DiffPhys, PhysNet-XY, and PhysNet-UV, we use the test results directly from the respective papers, which result in the absence of some experimental metrics. The remaining models are implement within the rPPG-Toolbox framework, ensuring consistent training and testing procedures for a fair comparison.} 

To evaluate the generalization ability of the rPPG model on out-of-distribution (OoD) data, we conduct cross-dataset testing on other three datasets after training on PURE or UBFC-rPPG. With the same settings of the rPPG-Toolbox (\cite{toolbox}),  we choose the UBFC-rPPG and PURE datasets for training because UBFC-Phys and MMPD datasets include lots of human motion and noise, which hinders model’s convergence. Specifically, the UBFC-Phys dataset involves considerable unconstrained motion under static lighting conditions, while the MMPD dataset includes data from subjects speaking, walking, and exercising. 

To quantitatively assess the predictive accuracy of various models, post-processing is applied to the outputted pulse wave signals. Specifically, the predicted waveforms are first filtered using a second-order Butterworth filter with cutoff frequencies of 0.75 and 2.5 Hz. Subsequently, the Fast Fourier Transform (FFT) is applied to the filtered signals to calculate the heart rates (HRs). The commonly used Mean Absolute Error (MAE), Mean Absolute Percentage Error (MAPE) and Pearson Correlation ($\rho$) are adopted as evaluation metrics:
\begin{equation}
    \operatorname{MAE}=\frac{1}{N} \sum_{n=1}^{N}\left|R_{g}-R_{p}\right|
\end{equation}
\begin{equation}
    \operatorname{MAPE}=\frac{1}{N} \sum_{n=1}^{N}\left|\frac{R_{g}-R_{p}}{R_{g}}\right|
\end{equation}
\begin{equation}
    \rho=\frac{\sum_{n=1}^{N}\left(R_{g . n}-\overline{R_{g}}\right)\left(R_{p . n}-\overline{R_{p}}\right)}{\sqrt{\left(\sum_{n=1}^{N} R_{g . n}-\overline{R_{g}}\right)^{2}\left(\sum_{n=1}^{N} R_{p . n}-\overline{R_{p}}\right)^{2}}}
\end{equation}
Where $R_p$, $R_g$, and $N$ denote the predicted signal rate, the ground truth signal rate, and the number of instances, respectively. Additionally, $\overline{R}$ is the average value of $R$ across $N$ samples. 

\subsection{Implementation details}
\textcolor{purple}{The proposed Spiking-PhysFormer is implemented with Pytorch and on two open-source platforms, SpikingJelly (\cite{b41}, \url{https://github.com/fangwei123456/spikingjelly}) and rPPG-Toolbox (\cite{toolbox}, \url{https://github.com/ubicomplab/rPPG-Toolbox})}, to facilitate fair comparisons. All videos in the datasets are resized to $128 \times 128$ by post-preprocessing, and a sequence of 160 consecutive frames is randomly selected as input, denoted as $\hat{X} \in \mathbb{R}^{3 \times 160 \times 128 \times 128}$. Additionally, we set the number of channels, \( D \), in the S3A to 96. We configure the timesteps $T_s$ of the SNN module to  4 and utilize four parallel spike-drive transformer blocks in Spiking-PhysFormer. The LIF neuron model's surrogate gradient function is $g(x)=\frac{1}{\pi}{\rm{arctan}}(\frac{1}{\pi}\alpha x)+\frac{1}{2}$, and its derivative is $g'(x)=\frac{\alpha}{2(1+(\frac{\pi}{2}\alpha x)^2)}$, where $\alpha$ represents the slope parameter. For all neurons, $\alpha=2$, $V_{reset}=0$, and $V_{th}=1$. Our model is trained with Adam optimizer and the initial learning rate and weight decay are 3e-3 and 5e-5, respectively. During the training phase, we first train the PhysFormer (\cite{physformer}) for 10 epochs using a standard configuration in rPPG-Toolbox (\cite{toolbox}) to obtain a pre-trained ANN-based PE block. Then, we train the Spiking-PhysFormer for 10 epochs, employing the model that performs the best on the validation set for cross-dataset testing. The batch size is set to 4, and the experiments are conducted on GeForce RTX 4090. In practice, we reshape the $Q, K, V \in \mathbb{R}^{T_s \times N \times D}$ (after Vid2Seq (\cite{physformer})) into multi-head form $q_{i}, k_i, v_i \in \mathbb{R}^{T_s \times N \times d}, i \in [1,..,H]$, where $D=$ $H \times d$. Next, we split $Q, K, V$ into $H$ parts and run $H$ S3A operations, in parallel, which are called multi-head S3A (MHS3A):  
\begin{equation}
    Q=\left(q_{1}, q_{2}, \cdots, q_{H}\right), K=\left(k_{1}, k_{2}, \cdots, k_{H}\right), V=\left(v_{1}, v_{2}, \cdots, v_{H}\right) 
\end{equation} 
\begin{equation}
    \operatorname{MHS3A'}(Q, K, V)=\left[\operatorname{S3A'}_{1}\left(q_{1}, k_{1}, v_{1}\right) ; \cdots ; \operatorname{S3A'}_{h}\left(q_{H}, k_{H}, v_{H}\right)\right]
\end{equation}
\begin{equation}
    \operatorname{MHS3A}(Q, K, V)=\operatorname{Seq2Vid}(\operatorname{BN}\left(\operatorname{Conv2D}\left(\mathcal{SN}\left(\operatorname{MHS3A'}(Q, K, V)\right)\right)\right))
\end{equation}

\subsection{Energy consumption analysis}
\definecolor{BlackPearl}{rgb}{0.003,0.015,0.035}

\begin{table}
\caption{\addtexttwo{Computational complexity comparisons of rPPG methods on frames with size of $128 \times 128$ (The FLOPs and Params of ANN-based models are derived from PhysBench (\cite{physbench}) and adjusted based on input size). The FLOPs for Spiking-PhysFormer are contributed by the ANN components, while the SOPs are from the SNN components.}}
\label{energy}
\centering
\begin{threeparttable}
\resizebox{\linewidth}{!}{
\begin{tblr}{
  cell{2}{2} = {r=6}{},
  cell{2}{3} = {r=6}{},
  hline{1-2,8-9} = {-}{},
}
Model                       & Method & Input size & Params & Frame FLOPs & Frame SOPs       & Power/mJ \\
TS-CAN                      & ANN    & 128x128    & 532K   & 717 M       & \textbackslash{} & 3.30     \\
PhysNet                     &        &            & 770K   & 864 M       & \textbackslash{} & 3.97     \\
DeepPhys                    &        &            & 532K   & 717 M       & \textbackslash{} & 3.30     \\
Eff.Phys-C                  &        &            & 2.16M  & 727 M       & \textbackslash{} & 3.34     \\
PhysFormer                  &        &            & 7.03M  & 324 M       & \textbackslash{} & 1.49     \\
iBVPNet                    &        &            & 1.43M  & 868 M       & \textbackslash{} & 3.99     \\
\textbf{Spiking-PhysFormer} & HNN    & 128x128    & 2.99M  & 290 M*       & 3.7 M            & 1.34     
\end{tblr}
}
 \begin{tablenotes}
        \footnotesize
        \item *The FLOPs of Spiking-PhysFormer are calculated solely for the ANN-based PE block and the predictor head.
\end{tablenotes}
\end{threeparttable}
\end{table}

\begin{figure}
\centering
\includegraphics[width=0.75\textwidth]{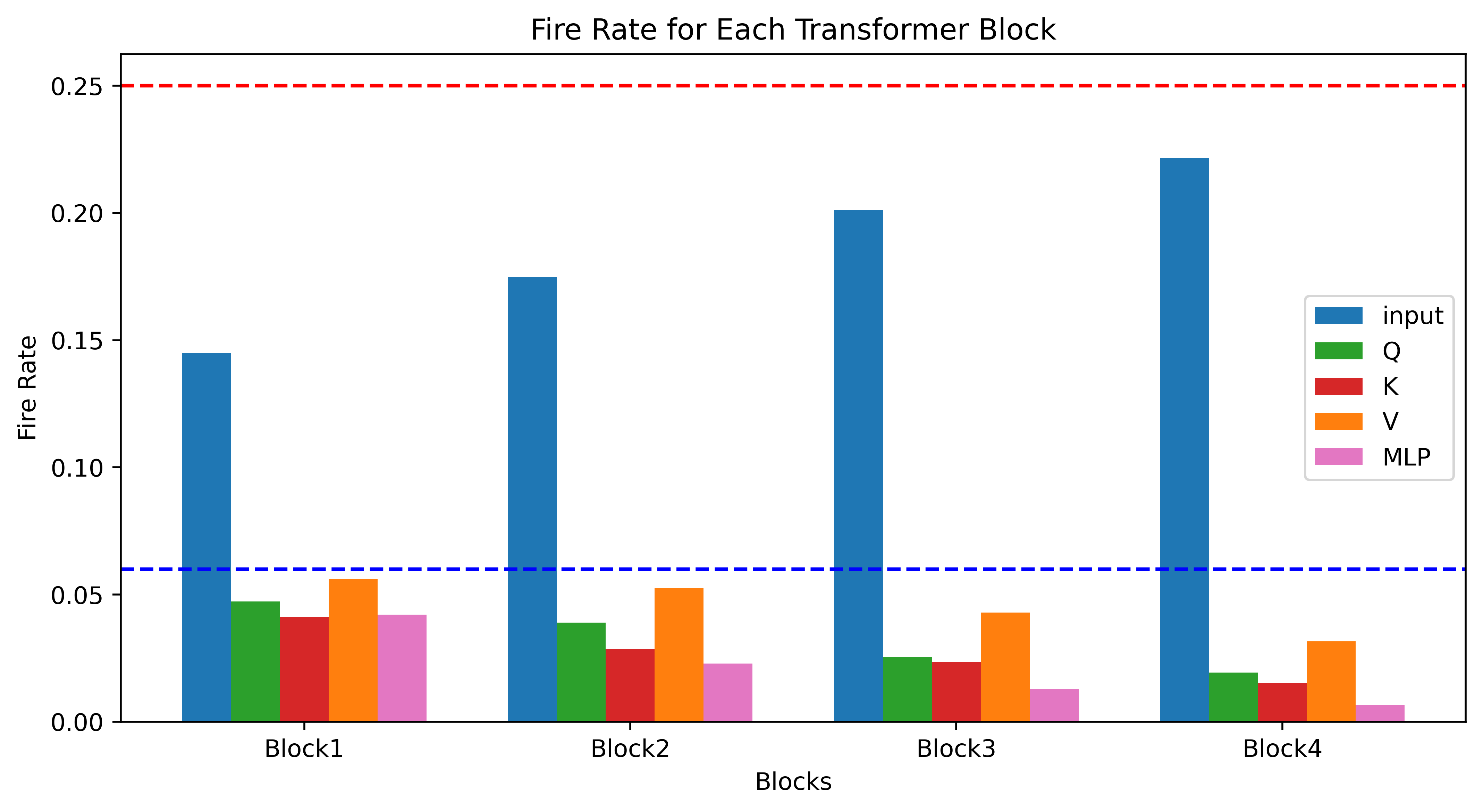}
\caption{Fire rate of input, $Q$, $K$, $V$ and MLP layer of transformer blocks in Spiking-PhysFormer.}
\label{fig8}
\end{figure}

\textcolor{purple}{The FLOPs of the ANN module in the proposed Spiking-PhysFormer can be conveniently obtained using the tool THOP (\url{https://github.com/Lyken17/pytorch-OpCounter}).} Now we need to analyze the energy consumption details of the SNN module. The FLOPs (the number of multiply-and-accumulate (MAC) operations) of ANN-based Conv3D layer ($\operatorname{FL}_{Conv3D}$) and Conv2D layer ($FL_{Conv2D}$) are:
\begin{equation}
    \operatorname{FLOPs}_{Conv3D}=\left(k_{n}\right)^{2} \cdot t_{n} \cdot h_{n} \cdot w_{n} \cdot c_{n-1} \cdot c_{n}
\end{equation}
\begin{equation}
    \operatorname{FLOPs}_{Conv2D}=\left(k_{n}\right)^{2} \cdot h_{n} \cdot w_{n} \cdot c_{n-1} \cdot c_{n}
\end{equation}
where $k_{n}$ is the kernel size, $\left(t_n, h_{n}, w_{n}\right)$ is the output feature map size, $c_{n-1}$ and $c_{n}$ are the input and output channel numbers, respectively. For the SNN-based convolution layer, the computation of theoretical energy consumption begins with the calculation of synaptic operations (SOPs) (\cite{spikformer}), which is the number of spike-based accumulate (AC) operations:
\begin{equation}
    \operatorname{SOPs}_{Conv2D} = fr \cdot T_s \cdot \operatorname{FLOPs}_{Conv2D} 
\end{equation}
\begin{equation}
    \operatorname{SOPs}_{Conv3D} = fr \cdot T_s \cdot \operatorname{FLOPs}_{Conv3D}
\end{equation}
where $fr$ and $T_s$ denote the spike fire rate and timesteps, respectively. \textcolor{purple}{The $fr$ is defined as the proportion of non-zero elements within the spike tensor. It is positively correlated with power consumption because when SNNs operate on neuromorphic hardware, sparse computations are triggered only when input spike signals arrive; otherwise, neurons remain quiescent.} Practically, we set $T_s$ to 4. Once the FLOPs for the ANN module and the SOPs for the SNN module are determined, we can further compute the energy cost $E$:
\begin{equation}
    E_{\operatorname{FLOPs}} = E_{\operatorname{MAC}} \times \operatorname{FLOPs}, \quad E_{\operatorname{SOPs}} = E_{\operatorname{AC}} \times \operatorname{SOPs}
\end{equation}
Drawing from previous studies (\cite{spike-drivenformer,spikformer}), we assume that the MAC and AC operations are implemented in 32-bit floating point format using 45nm technology (\cite{45nm}), where $E_{\operatorname{MAC}} = 4.6pJ$ and $E_{\operatorname{AC}} = 0.9pJ$. When $E_{\operatorname{AC}}\times T_s \times fr < E_{\operatorname{MAC}}$, SNNs are more energy-efficient compared to their ANN counterparts. Additionally, due to $T_s$ is a constant, the energy-saving characteristics of SNNs hinge on the spike firing rate $fr$.

Fig. \ref{fig8} illustrates the average spike fire rate, $fr$, for each tensor within the transformer blocks. We find that the input tensor's spike fire rate ($fr$) displays an ascending trend but consistently stays below 0.25 (as indicated by the red line in the Fig. \ref{fig8}), a consequence of the accumulative impact of inputs across layers, facilitated by residual connections. Conversely, the $fr$ of $Q$, $K$, and $V$ show a gradual decrease block by block, consistently staying below 0.06 (as indicated by the blue line in the Fig. \ref{fig8}), which indicates that the spatio-temporal attention mechanism guides the network to filter out less significant features as the network depth increases.

In conclusion, as shown in Table \ref{energy}, the computational complexity comparison reveals that the proposed Spiking-PhysFormer requires merely 1.34mJ to process a single-frame image of size $128 \times 128$. This represents a 10.1\% energy reduction compared to the PhysFormer and is substantially lower than that of other baseline models.

\subsection{Cross-dataset testing}

\begin{table}
\centering
\caption{Cross-dataset results training with the UBFC-rPPG dataset (\cite{ubfc-rppg}). $\operatorname{MAE}$ = Mean Absolute Error in HR estimation (Beats/Min), $\operatorname{MAPE}$ = Mean Percentage Error (\%), $\rho$ = Pearson Correlation in HR estimation. Best results are marked in \textcolor{red}{red}, second best in \textbf{bold}, and third best in \underline{underline}.}
\label{table1}
\resizebox{\linewidth}{!}{
\begin{tblr}{
  cell{1}{1} = {r=3}{},
  cell{1}{2} = {r=3}{},
  cell{1}{3} = {c=9}{},
  cell{2}{3} = {c=3}{},
  cell{2}{6} = {c=3}{},
  cell{2}{9} = {c=3}{},
  cell{4}{2} = {r=8}{},
  hline{1,4,12-13} = {-}{},
  hline{2-3} = {3-11}{},
}
Model         & Method & Test Set                         &                       &                                  &                  &                   &                   &                  &                   &                   \\
              &        & PURE                             &                       &                                  & UBFC-Phys        &                   &                   & MMPD             &                   &                   \\
              &        & MAE $\downarrow$                 & MAPE $\downarrow$     & $\rho$ $\uparrow$                & MAE $\downarrow$ & MAPE $\downarrow$ & $\rho$ $\uparrow$ & MAE $\downarrow$ & MAPE $\downarrow$ & $\rho$ $\uparrow$ \\
TS-CAN        & ANN    & \textcolor{red}{3.69}                             & \textcolor{red}{3.39}                  & 0.82                    & \textbf{5.13}    & \textbf{6.53}     & \textbf{0.76}     & 14.01            & 15.48             & \uline{0.24}     \\
PhysNet       &        & 8.06                             & 13.67                 & 0.61                             & \uline{5.79}     & \uline{7.69}      & \uline{0.70}      & \textbf{9.47}             & \textcolor{red}{11.11}             & \textbf{0.31}              \\
DeepPhys      &        & 5.54                             & \uline{5.32}         & 0.66                             & 6.62             & 8.21              & 0.66              & 17.50            & 19.27             & 0.06              \\
Eff.Phys-C    &        & 5.47                     & 5.40                  & 0.71                     & \textcolor{red}{4.93}             & \textcolor{red}{6.25}              & \textcolor{red}{0.79}              & 13.78    & \uline{15.15}    & 0.09              \\
PhysFormer    &        & 12.92                            & 23.92                 & 0.47                             & 6.63             & 8.91              & 0.69              & \uline{12.10}   & 15.41     & 0.17      \\
rFaceNet      &        & \textbf{3.74}  & \textbackslash{}      & \textcolor{red}{0.86}  & \textbackslash{} & \textbackslash{}  & \textbackslash{}  & \textbackslash{} & \textbackslash{}  & \textbackslash{}  \\
DiffPhys &        & 3.86~                            & \textbf{4.07}                  & \textbf{0.84}                             & \textbackslash{} & \textbackslash{}  & \textbackslash{}  & \textbackslash{} & \textbackslash{}  & \textbackslash{}  \\
iBVPNet       &        & 11.74                            & 21.99                 & 0.53                             & 5.82             & 7.73              & 0.66              & \textcolor{red}{9.23}             & \textbf{11.80}             & \textcolor{red}{0.35}              \\
\textbf{Ours} & HNN    & \uline{3.83$\pm$0.74}           & 5.70$\pm$0.74 & \uline{0.83$\pm$0.06}                    & 6.68$\pm$0.82    & 8.33$\pm$0.85     & 0.60$\pm$0.08     & 14.15$\pm$0.87   & 16.22$\pm$0.96    & 0.15$\pm$0.06     
\end{tblr}
}
\end{table}

\begin{figure}
\centering
\includegraphics[width=1\textwidth]{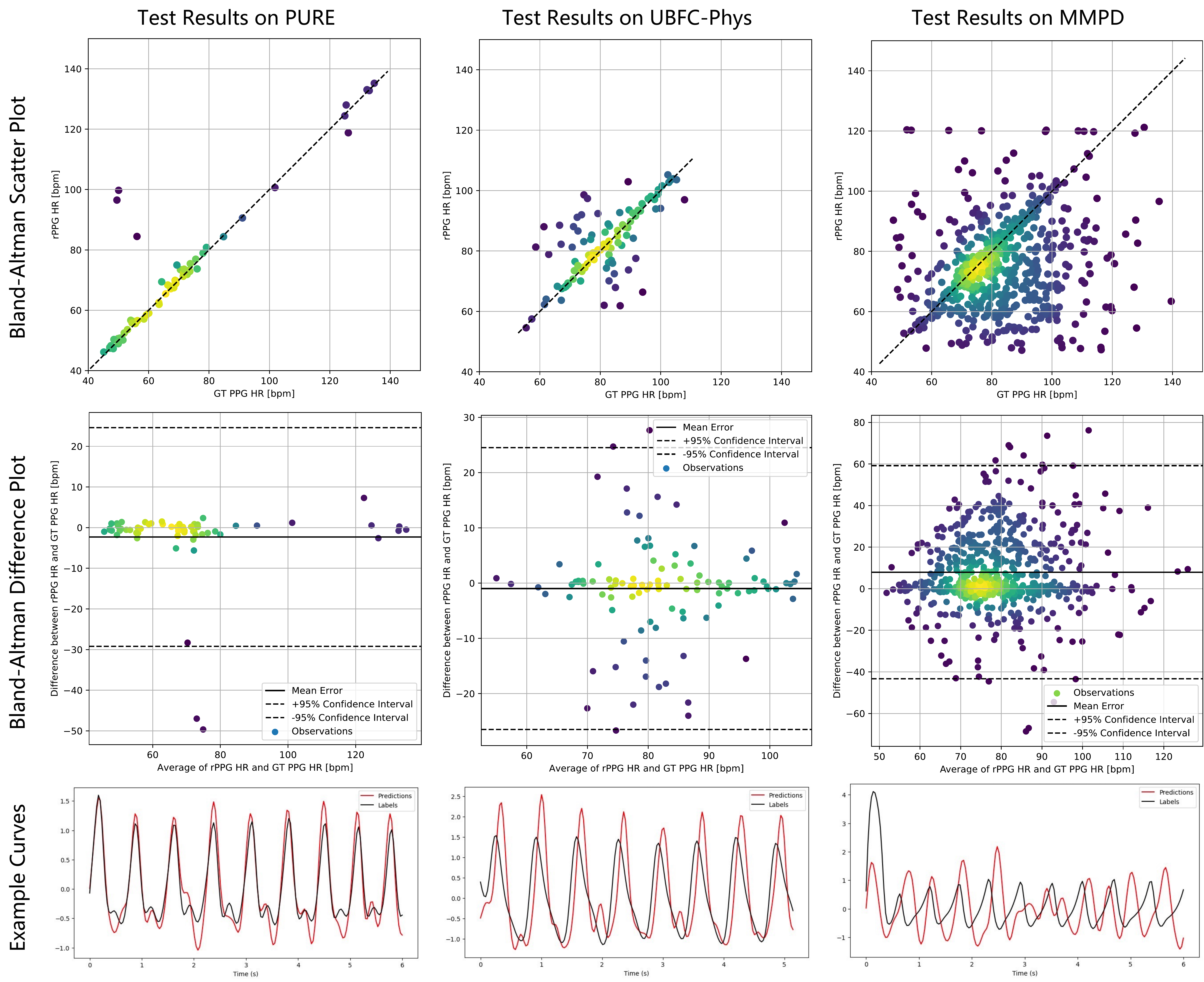}
\caption{Bland-Altman plots (\cite{baplots}) and output examples of cross-dataset results training with the UBFC-rPPG dataset (\cite{ubfc-rppg}).}
\label{fig6}
\end{figure}

\begin{table}
\centering
\caption{Cross-dataset results training with the PURE dataset (\cite{pure}). $\operatorname{MAE}$ = Mean Absolute Error in HR estimation (Beats/Min), $\operatorname{MAPE}$ = Mean Percentage Error (\%), $\rho$ = Pearson Correlation in HR estimation. Best results are marked in \textcolor{red}{red}, second best in \textbf{bold}, and third best in \underline{underline}.}
\label{table2}
\resizebox{\linewidth}{!}{
\begin{tblr}{
  cell{1}{1} = {r=3}{},
  cell{1}{2} = {r=3}{},
  cell{1}{3} = {c=9}{},
  cell{2}{3} = {c=3}{},
  cell{2}{6} = {c=3}{},
  cell{2}{9} = {c=3}{},
  cell{4}{2} = {r=9}{},
  cell{4}{10} = {fg=red},
  cell{5}{3} = {fg=red},
  cell{5}{4} = {fg=red},
  cell{5}{5} = {fg=red},
  cell{5}{7} = {fg=red},
  cell{5}{8} = {fg=red},
  cell{11}{9} = {fg=red},
  cell{11}{11} = {fg=red},
  cell{12}{6} = {fg=red},
  hline{1,4,13-14} = {-}{},
  hline{2-3} = {3-11}{},
}
Model         & Method & Test Set         &                   &                   &                  &                       &                   &                  &                   &                   \\
              &        & UBFC-rPPG        &                   &                   & UBFC-Phys        &                       &                   & MMPD             &                   &                   \\
              &        & MAE $\downarrow$ & MAPE $\downarrow$ & $\rho$ $\uparrow$ & MAE $\downarrow$ & MAPE $\downarrow$     & $\rho$ $\uparrow$ & MAE $\downarrow$ & MAPE $\downarrow$ & $\rho$ $\uparrow$ \\
TS-CAN        & ANN    & \uline{1.30}     & \uline{1.50}      & \uline{0.99}      & 5.72             & 7.34                  & \textbf{0.72}     & \uline{13.94}    & 15.14             & \textbf{0.20}     \\
PhysNet       &        & 0.98             & 1.12              & 0.99              & \textbf{4.78}    & 6.15                  & 0.73              & \textbf{13.93}   & \uline{15.61}     & \uline{0.17}      \\
DeepPhys      &        & 1.21    & \textbf{1.42}     & 0.99     & 8.42             & 10.18                 & 0.44              & 16.92            & 18.54             & 0.05              \\
Eff.Phys-C    &        & 2.07             & 2.10              & 0.94              & \uline{5.31}     & \textbf{6.61}         & \uline{0.70}      & 14.03            & \textbf{15.31}    & 0.17              \\
PhysFormer    &        & 1.44             & 1.66              & 0.98              & 6.04             & 7.67                  & 0.65              & 14.57            & 16.73             & 0.15              \\
rFaceNet      &        & \textbf{1.05}    & \textbackslash{}  & \textbf{0.99}              & \textbackslash{} & \textbackslash{}      & \textbackslash{}  & \textbackslash{} & \textbackslash{}  & \textbackslash{}  \\
PhysNet-XY    &        & \textbackslash{} & \textbackslash{}  & \textbackslash{}  & \textbackslash{} & \textbackslash{}      & \textbackslash{}  & 14.91            & \textbackslash{}  & 0.15              \\
PhysNet-UV    &        & \textbackslash{} & \textbackslash{}  & \textbackslash{}  & \textbackslash{} & \textbackslash{}      & \textbackslash{}  & 12.19            & \textbackslash{}  & 0.29              \\
iBVPNet       &        & 5.21             & 5.10              & 0.83              & 4.51             & 7.88                  & 5.80              & 15.78            & 17.06             & 0.04              \\
\textbf{Ours} & HNN    & 2.80$\pm$1.69    & 2.81$\pm$1.46     & 0.95$\pm$0.06     & 5.72$\pm$0.75    & \uline{7.17$\pm$0.79} & 0.68$\pm$0.07     & 14.57$\pm$1.52   & 16.55$\pm$1.34    & 0.14$\pm$0.06     
\end{tblr}
}
\end{table}
\begin{figure}
\centering
\includegraphics[width=1\textwidth]{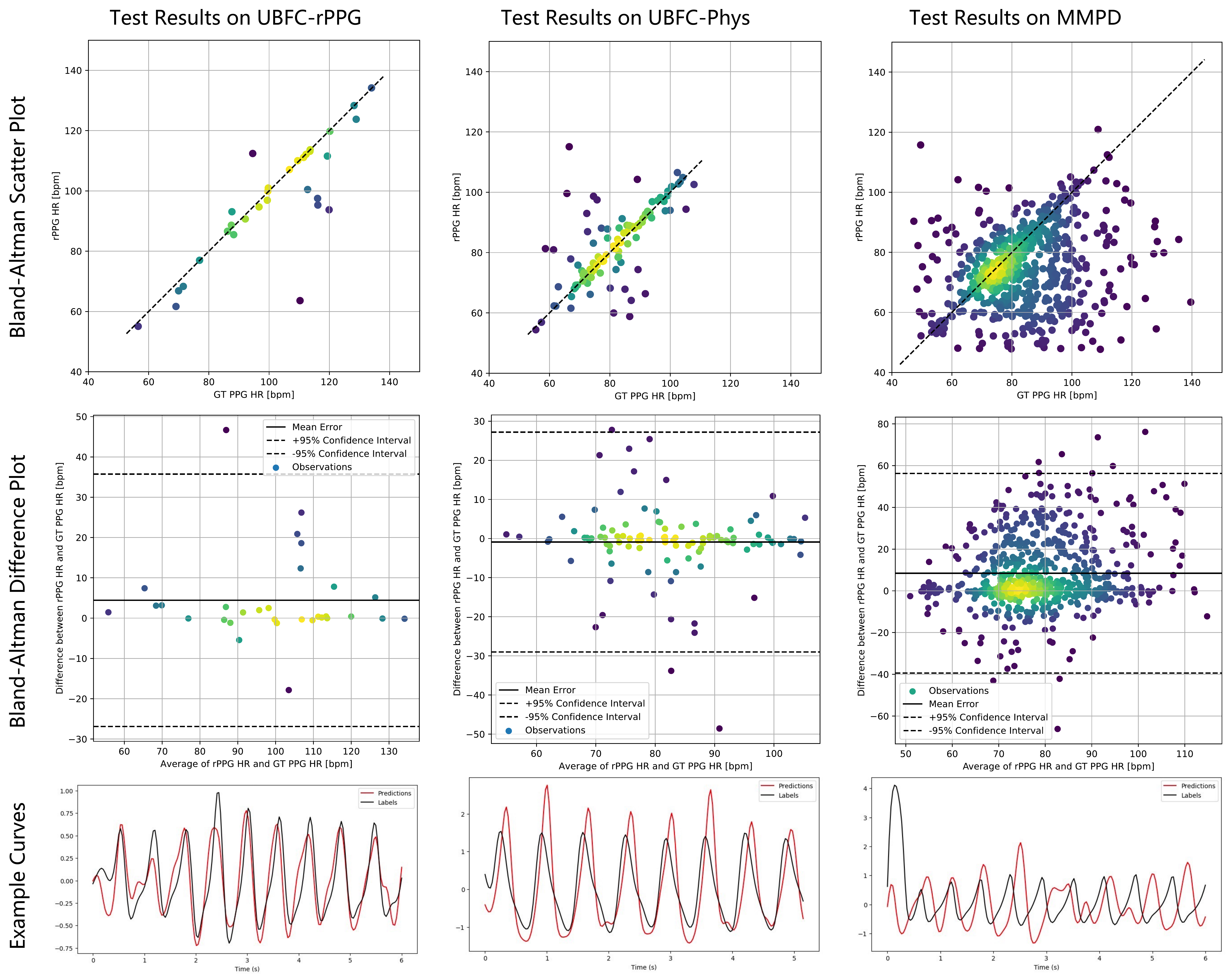}
\caption{Bland-Altman plots (\cite{baplots}) and output examples of cross-dataset results training with the PURE dataset (\cite{pure}).}
\label{fig7}
\end{figure}

Performance metrics, detailed in Table \ref{table1} and Fig. \ref{fig6}, are obtained by training on UBFC-rPPG and testing on three other datasets. Similarly, metrics in Table \ref{table2} and Fig. \ref{fig7} are derived from training on PURE and subsequent testing on three other datasets. This cross-dataset testing verifies the model's adaptability to videos with diverse facial features, backgrounds, and illumination.

\textbf{Results from training on UBFC-rPPG}: \textcolor{purple}{As shown in Table \ref{table1}, Spiking PhysFormer achieves performance comparable to the current state-of-the-art ANN-based methods on all three test datasets.} Additionally, we can see that Spiking-PhysFormer significantly outperforms PhysFormer on the PURE dataset (MAE: 12.92 $\rightarrow$ 3.83), and exhibits comparable performance on UBFC-Phys and MMPD. This improvement can be attributed to the enhanced capability of the proposed spike-driven S3A to extract pivotal features, and the long-range spatio-temporal attention mechanism, which mitigates overfitting on in-distribution (ID) data. To further examine the correlations between the predicted HRs and the ground-truth HRs, we present Bland-Altman plots (\cite{baplots}) in Fig. \ref{fig6}. As demonstrated in the top two rows, Spiking-PhysFormer shows a strong correlation with the ground-truth HRs across a wide HR range of 40 to 140 bpm. Moreover, as observed from the output examples in the third row, the predictive challenge on the MMPD is considerably higher than on PURE and UBFC-Phys, which is due to the diversity in the MMPD dataset, where subjects engage in various types of physical activities and possess a range of skin tones, complicating the generalization of models pre-trained on UBFC-rPPG to this dataset.

\textbf{Results from training on PURE}: \textcolor{purple}{The Spiking PhysFormer demonstrates comparable predictive performance to the PhysFormer on MMPD, ranking third on UBFC Phys and outperforming PhysFormers in terms of performance.} However, we find that the test performance on UBFC-rPPG is inferior to that of the ANN-based model, which may be attributed to the limited sample size of UBFC-rPPG, resulting in individual sample test errors having a disproportionate impact on the overall metrics. The Bland-Altman plot and output examples for UBFC-rPPG depicted in Fig. \ref{fig7} reveal that Spiking-PhysFormer's predictions are consistent with the ground-truth HRs across the majority of samples.

To sum up, the proposed Spiking-PhysFormer demonstrates equivalent performance to the state-of-the-art ANN-based rPPG models with less power consumption (Section 4.3), which indicates that the Spiking-PhysFormer can balance efficiency and accuracy.

\subsection{Spatio-temporal attention map}
\begin{figure}[htbp]
\centering
\includegraphics[width=0.7\textwidth]{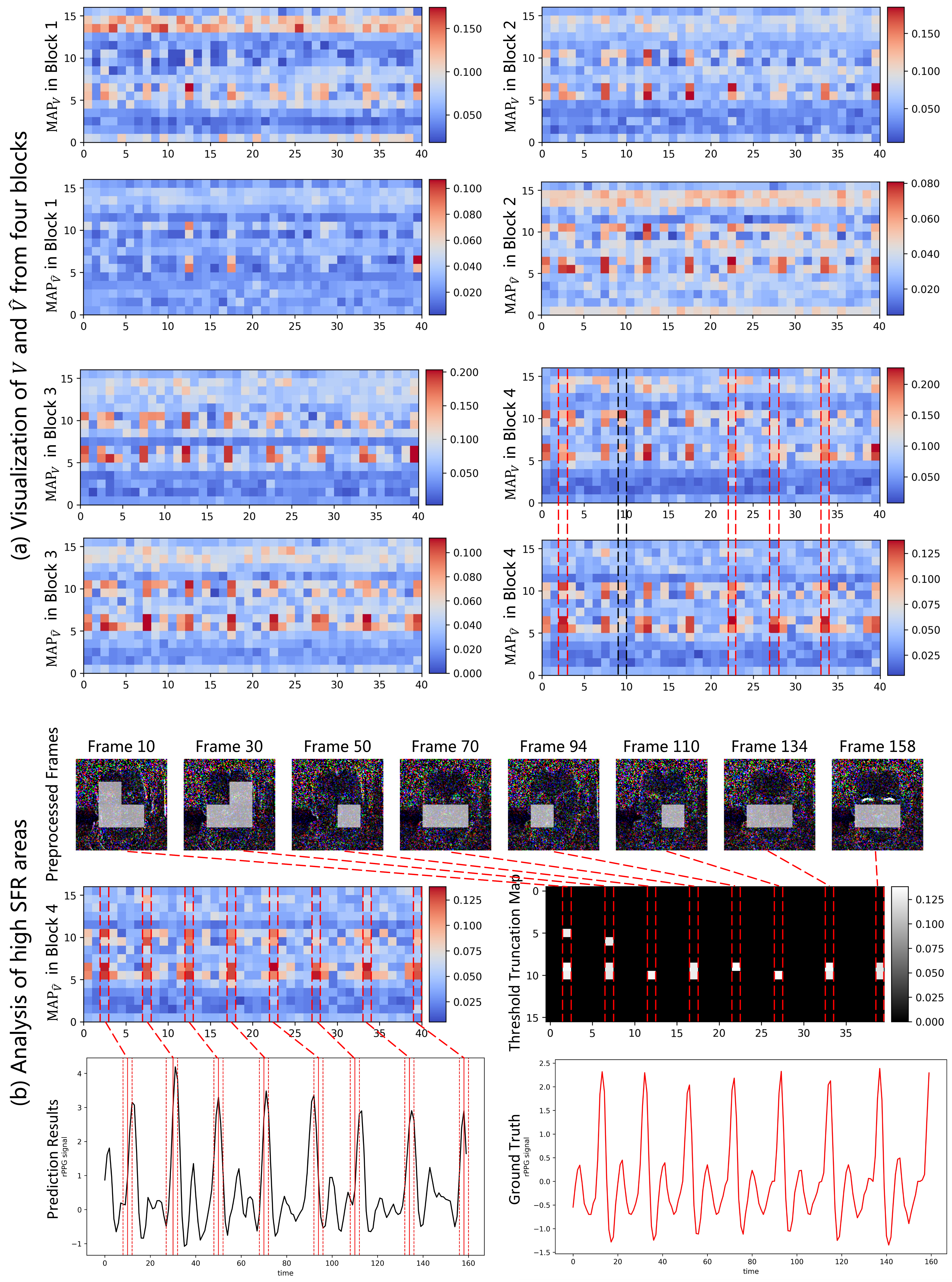}
\caption{Spatio-temporal attention map based on spike firing rate (SFR), the redder the higher the SFR, the bluer the smaller the SFR. (a) The visualization of $V$ and $\hat{V}$ from the four parallel spike-driven transformer blocks reveals that irrelevant background signals and noise in $V$ are masked by the hard attention mechanism. Consequently, $\hat{V}$ extracted contains critical spatio-temporal features. Furthermore, from the first block to the fourth, the key features within $\hat{V}$ become progressively more distinct. (b) Analysis of high SFR areas. By overlaying the high SFR areas from the threshold truncation map onto the input image, we find that the S3A mechanism directs the model's focus to the facial regions within the image. Moreover, it is observable that the areas with high SFR correspond to the peaks of the pulse wave in the temporal domain.}
\label{fig9}
\end{figure}

As described in Section 3.3, upon obtaining $V$, $Q$, and $K$ $\in \mathbb{R}^{T_s \times N \times D}$, the proposed S3A is:
\begin{equation}
    \hat{V} = \operatorname{S3A'}(Q, K, V) = g(Q, K) \otimes V = \mathcal{SN}((Q \otimes K)) \otimes V
\end{equation}
\begin{equation}
    \operatorname{S3A}(Q, K, V) = \operatorname{Seq2Vid}(\operatorname{BN}(\operatorname{Conv2D}(\mathcal{SN}(\hat{V}))))
\end{equation}
We define the output of $\operatorname{S3A'}(Q, K, V)$ as $\hat{V} \in \mathbb{R}^{T_s \times N \times D}$. Given that the output from $g(Q,K) = \mathcal{SN}(\operatorname{SUM}_c(Q \otimes K))$ is a D-dimensional row vector composed exclusively of binary spike sequences, the S3A we propose qualifies as a hard attention mechanism (\cite{hardatten}). S3A effectively masks the non-significant channels within V, ensuring that $\hat{V}$ retains only the essential spatio-temporal features, which guide the downstream modules for precise pulse wave prediction.

To demonstrate how the hard attention mechanism within S3A modulates the SFR of $V$, we present the visualization results in Fig. \ref{fig9}. To obtain the spatio-temporal attention map based on SFR, we first compute the SFR by averaging the spike and channel dimensions of $V$ and $\hat{V}$. Subsequently, we unfold these averages across temporal and spatial dimensions to derive $\operatorname{MAP}_V$ and $\operatorname{MAP}_{\hat{V}} \in \mathbb{R}^{\hat{T} \times \hat{H}\hat{W}}$, where $N = \hat{T} \times \hat{H}\hat{W}$. 

As depicted in Fig. \ref{fig9}(a), we observe a progressive enhancement of high SFR regions within $\operatorname{MAP}_{\hat{V}}$ from shallower to deeper blocks, indicating a refinement of key features. In the fourth block, highlighted by the red dashed lines, some low SFR areas in $\operatorname{MAP}_V$ become relatively high SFR regions in $\operatorname{MAP}_{\hat{V}}$ after masking by the hard attention S3A mechanism. Conversely, the black dashed lines illustrate the opposite effect. Above findings confirm that the attention scores can modulate the spike firing in $V$.

To further validate the interpretability of the spatio-temporal attention in S3A, we correlated the high SFR regions in $\operatorname{MAP}_{\hat{V}}$ with the input frames and the output pulse wave signals. The results are depicted in Fig. \ref{fig9}(b), where the threshold truncation map is the result of truncating $\operatorname{MAP}_{\hat{V}}$ at a threshold of 0.1, preserving only the high SFR regions. When these regions are superimposed on the corresponding input frames, it is apparent that the high SFR areas are confined to the facial region. This feature enables the model to autonomously focus on changes in facial features without the need of an ROI mask. Additionally, Fig. \ref{fig9}(b) reveals that the high SFR regions correspond temporally with the peak values of the pulse wave, indicating that the attention-score-masked $\hat{V}$ can guide the downstream model to accurately predict the pulse wave and HR.

\subsection{Ablation study}
To demonstrate the effectiveness of Spiking-PhysFormer, we carry out ablation studies on different factors such as the number of transformer blocks, the impact of parallel sub-blocks, self-attention methods and the ANN components. \textcolor{purple}{All experiments were conducted with fixed random seed for training and testing, ensuring reproducibility across single runs.}

\textbf{Impact of block numbers.} 
\begin{table}
\centering
\caption{Impact of the number of transformer blocks on PURE dataset, PhysFormer and Spiking-PhysFormer are trained on UBFC-rPPG. Best results are marked in \textcolor{red}{red}.}
\label{ablation_block}
\resizebox{0.7\linewidth}{!}{
\centering
\begin{tblr}{
  cell{1}{1} = {r=2}{},
  cell{1}{2} = {r=2}{},
  cell{1}{3} = {c=3}{},
  cell{3}{1} = {r=5}{},
  cell{8}{1} = {r=5}{},
  cell{11}{3} = {fg=red},
  cell{11}{4} = {fg=red},
  cell{11}{5} = {fg=red},
  hline{1,3,8,13} = {-}{},
  hline{2} = {3-5}{},
}
Model                       & Blocks & Test on PURE     &                   &                   \\
                            &        & MAE $\downarrow$ & MAPE $\downarrow$ & $\rho$ $\uparrow$ \\
PhysFormer (\cite{physformer})               & 4      & 10.39            & 22.08             & 0.49              \\
                            & 6      & 9.12             & 15.61             & 0.60              \\
                            & 8      & 10.2             & 18.53             & 0.60              \\
                            & 10     & 11.05            & 18.52             & 0.49              \\
                            & 12     & 12.92            & 23.92             & 0.47              \\
\textbf{Spiking-PhysFormer} & 4      & 3.32             & 4.91              & 0.88              \\
                            & 6      & 3.72             & 7.36              & 0.89              \\
                            & 8      & 5.79             & 8.71              & 0.72              \\
                            & 10     & 2.90             & 5.47              & 0.91              \\
                            & 12     & 4.16             & 8.28              & 0.86              
\end{tblr}
}
\end{table}
As illustrated in Table \ref{ablation_block}, we investigate the impact of the number of transformer blocks on model performance. Given that PhysFormer (\cite{physformer}) is also a transformer-based rPPG model, we conduct a comparative study. Contrary to the conclusions of most transformer-based research (\cite{physformer, single}), we find that for rPPG tasks, both PhysFormer and Spiking-PhysFormer models yield better performance with fewer transformer blocks. Specifically, PhysFormer achieves optimal performance with 6 layers, while performance is at its lowest with 12 layers, contradicting the findings of ablation studies by \cite{physformer}, who suggested deeper transformer blocks for better performance. The discrepancy arises as we use cross-data testing to assess the models' generalization capabilities on OoD data, where an excessive number of transformer blocks can lead to overfitting. In contrast, \cite{physformer} tests their model on ID data. For the Spiking-PhysFormer, the best results are achieved with 10 layers; however, using 4 or 6 layers also yields comparable results (2.90 vs 3.32 or 3.72). Therefore, for rPPG tasks, fewer block layers can alse achieve generalization performance and significantly reduce the number of parameters.

\textbf{Impact of timesteps of the SNN module.}
\begin{table}
\centering
\caption{Impact of timesteps of the SNN module on PURE dataset, Spiking-PhysFormer is trained on UBFC-rPPG. Best results are marked in \textcolor{red}{red}.}
\label{ablation_steps}
\resizebox{0.7\linewidth}{!}{
\begin{tblr}{
  cell{1}{1} = {r=2}{},
  cell{1}{2} = {r=2}{},
  cell{1}{3} = {r=2}{},
  cell{1}{4} = {c=3}{},
  cell{3}{1} = {r=5}{},
  hline{1,3,8} = {-}{},
  hline{2} = {4-6}{},
}
Model              & timesteps & Power (mJ) & Test on PURE &       &      \\
                   &           &            & MAE $\downarrow$         & MAPE $\downarrow$  & rho $\uparrow$  \\
\textbf{Spiking-PhysFormer} & 1         & 1.334      & \textcolor{red}{2.56}         & \textcolor{red}{4.13}  & \textcolor{red}{0.92} \\
                   & 2         & 1.335      & 4.22         & 6.09  & 0.82 \\
                   & 4         & 1.337      & 3.32          & 4.91  & 0.88 \\
                   & 8         & 1.340       & 2.72         & 5.23  & 0.92 \\
                   & 16        & 1.347      & 7.41         & 14.54 & 0.76 
\end{tblr}
}
\end{table}
Timesteps play a crucial role in SNNs. Previous research indicates that larger timesteps can enhance the performance of SNNs, as longer spike sequences are capable of encoding more information (\cite{spike-drivenformer,TransformerbasedSN,spikingddpm,spikingdiffusion}). However, the proposed Spiking-PhysFormer is an HNN, in which the SNN module is engineered to direct the model's attention to crucial spatio-temporal features, thus fewer timesteps actually benefit the hard attention mechanism in filtering out background noise. The ablation study presented in Table \ref{ablation_steps} shows that Spiking-PhysFormer's performance remains stable with timesteps set to 1, 2, 4, and 8. A notable decrease in performance occurs when the timestep is increased to 16, as excessively long spike sequences introduce redundant information that may hinder the attention mechanism.

\begin{table}
\centering
\caption{Impact of parallel sub-blocks on PURE dataset, PhysFormer and Spiking-PhysFormer are trained on UBFC-rPPG. Best results are marked in \textcolor{red}{red}.}
\label{ablation_parallel}
\resizebox{0.7\linewidth}{!}{
\begin{tblr}{
  cell{1}{1} = {r=2}{},
  cell{1}{2} = {r=2}{},
  cell{1}{3} = {c=3}{},
  cell{3}{1} = {r=2}{},
  cell{5}{1} = {r=2}{},
  hline{1,3,7} = {-}{},
  hline{2} = {3-5}{},
  hline{5} = {2-5}{},
}
Model                       & sub-blocks   & Test on PURE     &                   &                   \\
                            &              & MAE $\downarrow$ & MAPE $\downarrow$ & $\rho$ $\uparrow$ \\
PhysFormer (\cite{physformer})               & w/o parallel & 12.92            & 23.92             & 0.47              \\
                            & parallel     & 9.90             & 17.66             & 0.68              \\
\textbf{Spiking-PhysFormer} & w/o parallel & 5.30             & 10.49             & 0.83              \\
                            & parallel     & \textcolor{red}{3.32}             & \textcolor{red}{4.91}              & \textcolor{red}{0.88}              
\end{tblr}
}
\end{table}

\begin{figure}[htbp]
\centering
\includegraphics[width=1\textwidth]{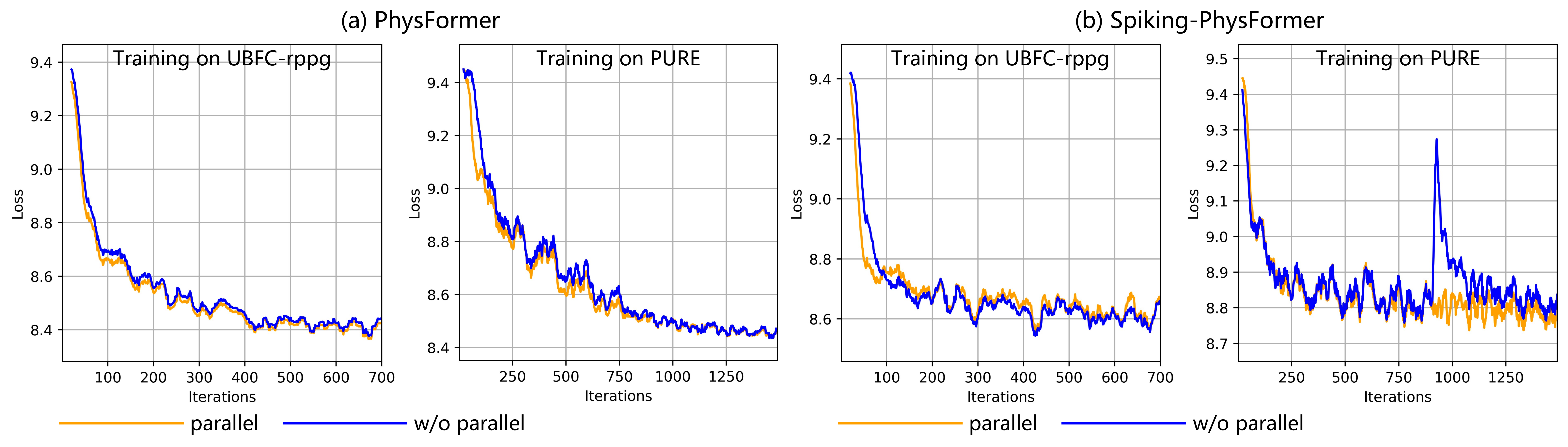}
\caption{\addtexttwo{Impact of parallel sub-blocks on the training convergence of PhysFormer and Spiking-PhysFormer}}
\label{fig10}
\end{figure}

\textbf{Impact of parallel sub-blocks.} Parallelizing sub-blocks within the transformer block facilitates faster inference speeds on hardware (\cite{gpt-j,simplifying}). Additionally, the proposed SFR-based attention map, which provides essential spatio-temporal features, is beneficial for predicting pulse wave signals when directly fed to the subsequent head predictor. As indicated in Table \ref{ablation_parallel}, our ablation study investigates the effects of configuring sub-blocks in PhysFormer and Spiking-PhysFormer to operate in parallel. The results demonstrate that for PhysFormer, parallel sub-blocks significantly enhance performance on the PURE dataset (MAE reduced from 12.92 to 9.90). For Spiking-PhysFormer, transitioning to a parallel arrangement improves results on PURE dataset. In summary, for rPPG tasks, parallel sub-blocks offer enhanced model generalizability and faster inference speeds. \addtexttwo{To further assess how parallelization affects training stability, we plotted the training loss for Spiking-PhysFormer and PhysFormer using both serial and parallel transformer blocks. As shown in Fig. \ref{fig10}, parallel sub-blocks improved the training stability of both models. More concretely, PhysFormer demonstrates faster convergence, while for Spiking-PhysFormer, training on the PURE dataset with serial blocks led to unstable fluctuations in loss values, manifesting as the loss spike (\cite{li2023loss}). In contrast, parallel blocks effectively mitigated this issue. Furthermore, although Spiking-PhysFormer exhibits a higher loss on the training set upon completion of training, it demonstrates comparable or superior performance in cross-dataset testing. This is particularly evident when trained on the UBFC-rPPG dataset and tested on the PURE dataset (12.92 vs 3.83), indicating that Spiking-PhysFormer effectively mitigates overfitting.}

\textbf{Impact of the attention block.}
\begin{table}
\centering
\caption{Impact of the attention block on PURE dataset, PhysFormer and Spiking-PhysFormer are trained on UBFC-rPPG. Best results are marked in \textcolor{red}{red}.}
\label{ablation_attention}
\begin{threeparttable}
\resizebox{\linewidth}{!}{
\begin{tblr}{
  cell{1}{1} = {r=2}{},
  cell{1}{2} = {c=3}{},
  cell{1}{5} = {r=2}{},
  cell{1}{6} = {c=3}{},
  cell{4}{1} = {r=4}{},
  hline{1,3-4,8} = {-}{},
  hline{2} = {2-4,6-8}{},
}
Model              & Projection layers &        &        & Power*($\mu J$) & Test on PURE &       &      \\
                   & $Q$                & $K$      & $V$      &            & MAE $\downarrow$          & MAPE $\downarrow$  & $\rho$ $\uparrow$ \\
PhysFormer (\cite{physformer})        & TDC               & TDC    & Conv3D & 10.13      & 12.92        & 23.92 & 0.47 \\
\textbf{Spiking-PhysFormer} & TDC               & TDC    & Conv3D & 1.52       & \textcolor{red}{2.49}         & \textcolor{red}{4.83}  & \textcolor{red}{0.92} \\
                   & TDC               & Conv3D & Conv3D & 0.85       & 4.42         & 8.50   & 0.87 \\
                   & \textbf{TDC}               & \textbf{Conv3D} & \textbf{None}   & 0.83       & 3.32          & 4.91  & 0.88 \\
                   & Conv3D            & Conv3D & None   & 0.16       & 4.77         & 7.15  & 0.80  
\end{tblr}
}
 \begin{tablenotes}
        \footnotesize
        \item *The power consumption required for a single transformer block.
\end{tablenotes}
\end{threeparttable}
\end{table}
To reduce computational overhead without compromising performance, we simplify the attention block. The results of the ablation study are presented in Table \ref{ablation_attention}. Because of the binary computation and sparsity in SNNs, the Spiking-PhysFormer's transformer block only needs 1.52 $\mu J$ of energy, marking a $6.7 \times$ decrease compared to the PhysFormer (\cite{physformer}), even when using the same projection layers, such as TDC for computing $Q$ and $K$, and a Conv3D layer for $V$. Besides, we simplify the attention layer by reducing the TDC layers, as the DiffNormalized preprocessing method is used to capture frame differences. The ablation study reveals that eliminating one TDC layer and the Conv3D layer used to compute $V$ results in only a minor performance degradation (MAE increases from 2.49 to 3.32) while reducing power consumption by 45.4\%. To sum up, we propose that adopting a TDC layer for computing $Q$, a Conv3D layer for $K$, and omitting the projection layer for $V$ represents a viable strategy to balance performance with power efficiency, as delineated in the S3A mechanism.

\addtexttwo{\textbf{Impact of the ANN components.} To illustrate that the combined use of ANN and SNN yields the best performance, we presented further findings from an ablation study, as detailed in Table 8. When trained on UBFC-rppg, the pure SNN model achieved a MAE of 12.39, markedly lower than the HNN, which builds the transformer block using SNN while relying on ANN for the PE block and predictor.}
\begin{table}
\centering
\caption{\addtexttwo{Impact of the ANN components, Spiking-PhysFormer are trained on UBFC-rPPG. Best results are marked in \textcolor{red}{red}.}}
\label{ablation_ann}
\resizebox{\linewidth}{!}{
\begin{tblr}{
  cell{1}{1} = {r=2}{},
  cell{1}{2} = {c=3}{},
  cell{1}{5} = {c=3}{},
  cell{3}{1} = {r=3}{},
  hline{1,3,6} = {-}{},
  hline{2} = {2-7}{},
}
Model                       & Component    &                   &                & Test on PURE     &                   &                   \\
                            & PE block     & Transformer block & Predictor head & MAE $\downarrow$ & MAPE $\downarrow$ & $\rho$ $\uparrow$ \\
\textbf{Spiking-PhysFormer} & SNN          & SNN               & SNN            & 12.39            & 23.05             & 0.50              \\
                            & ANN          & SNN               & SNN            & 9.56             & 12.02             & 0.43              \\
                            & \textbf{ANN} & \textbf{SNN}      & \textbf{ANN}   & \textcolor{red}{3.32}             & \textcolor{red}{4.91}              & \textcolor{red}{0.88}              
\end{tblr}
}
\end{table}

\addtext{
\section{Discussion}
\subsection{Limitations of the Study}
Although the proposed method in this paper balances performance and power consumption and has been comprehensively compared with existing methods to confirm its effectiveness, there are still some limitations. Firstly, our rPPG model is tested on a Linux platform, not in practical scenarios. Secondly, the model needs extensive testing across various conditions such as motion, skin tone, and lighting, which are critical for the algorithm’s robustness in diverse environments. Addressing these problems is crucial for the future development of rPPG technology and its successful application in various real-world scenarios.
\subsection{Impacts of the Study}
The advancement of rPPG technology holds promising potential for positive societal impacts, particularly in the context of remote healthcare and continuous physiological monitoring. As the technology matures, it is expected that rPPG can be integrated into endpoint devices, facilitating ongoing inference and health monitoring without interfering with the device's core functionalities. This integration could significantly reduce energy consumption during algorithm deployment, enhancing the feasibility of widespread use in medical and everyday settings. Such developments could revolutionize remote patient monitoring by providing real-time health data, thereby enabling timely medical interventions and promoting overall public health. Despite its potential benefits, the deployment of rPPG technology raises substantial concerns regarding privacy and ethical implications. One of the most pressing issues is the technology's ability to collect physiological data through monitoring cameras without the user's consent. This capability could lead to the leakage of sensitive personal health information, posing significant risks. Such data could be exploited by employers to assess employee productivity or even used as a basis for legal scrutiny. The potential for misuse of this sensitive information underscores the need for stringent ethical standards and robust privacy protections to be established as integral components of rPPG technology deployment.
}

\section{Conclusions}
In pursuit of a camera-based remote photoplethysmography (rPPG) solution that balances energy efficiency with performance, this paper introduces the Spiking-PhysFormer, a hybrid neural network (HNN) composed of an ANN-based patch embedding block and predictor head, alongside SNN-based parallel transformer blocks. To streamline the transformer block, we design a backbone consisting of parallel sub-blocks and introduce the S3A mechanism for extracting spatio-temporal key features, utilizing a single temporal difference convolution (TDC) layer to reduce the parameter count and computational expenditure. This marks the first incorporation of SNNs into the rPPG domain, injecting new perspectives and potential avenues of exploration. By generating spatio-temporal attention maps based on the spike firing rate (SFR), we validate the interpretability of our proposed model. Experiments across four datasets demonstrate that the Spiking-PhysFormer significantly reduces computational power consumption while maintaining predictive accuracy and generalization capabilities comparable to ANN-based models.

\section*{Acknowledgement}
This work is supported by the National Science and Technology Major Project from Minister of Science and Technology, China (Grant No. 2018AAA0103100), the National Natural Science Foundation of China (No. 92164110, No. 62132010, No. 62002198, and No. 62334014), and the Laboratory Open Fund  of Beijing Smartchip Microelectronics Technology Co., Ltd (No. SGTYHT/21-JS-223), partly supported by Beijing Engineering Research Center (No. BG0149), Young Elite Scientists Sponsorship Program by CAST under Grant No.2021QNRC001, and Beijing Natural Science Foundation (No.QY23124). The corresponding authors of this paper are Hong Chen and Yuntao Wang.

\section*{Declaration of generative AI and AI-assisted technologies in the writing process
}

During the preparation of this work the authors used GPT-4 in order to polish the content. After using this tool/service, the authors reviewed and edited the content as needed and take full responsibility for the content of the publication.

\bibliographystyle{elsarticle-harv} 
\bibliography{references}{}

\end{document}